\documentclass[10pt]{article} 

\usepackage[accepted]{rlj} 
\usepackage{amssymb}           
\usepackage{mathtools}         
\usepackage{mathrsfs}          
\usepackage{graphicx}         
\usepackage{subcaption}
\usepackage[space]{grffile}
\usepackage{url}
\usepackage{lipsum}

\renewcommand{\cite}{\citep}

\usepackage{nicefrac}
\usepackage{booktabs}
\usepackage{multirow}

\title{V-Max: A Reinforcement Learning Framework \\ for Autonomous Driving}

\setrunningtitle{V-Max: A RL Framework for Autonomous Driving}

\author{Valentin Charraut\textsuperscript{$1, \dagger$}, Waël Doulazmi\textsuperscript{$1,2, \dagger$}, Thomas Tournaire\textsuperscript{$1, \dagger$},
Thibault Buhet\textsuperscript{$1$}}

\emails{\{firstname.lastname\}@valeo.com}

\affiliations{
$1$ \textbf{Valeo Brain}\\
$2$ \textbf{Centre for Robotics, Mines Paris - PSL}\\
\par 
$^\dagger$ Equal contribution
}


\summary{Learning-based decision-making has the potential to enable generalizable Autonomous Driving (AD) policies, reducing the engineering overhead of rule-based approaches. Imitation Learning (IL) remains the dominant paradigm, benefiting from large-scale human demonstration datasets, but it suffers from inherent limitations such as distribution shift and imitation gaps. Reinforcement Learning (RL) presents a promising alternative, yet its adoption in AD remains limited due to the lack of standardized and efficient research frameworks. To this end, we introduce V-Max, an open research framework providing all the necessary tools to facilitate RL research for AD. V-Max is built on Waymax \cite{gulino_waymax_2023}, a hardware-accelerated AD simulator designed for large-scale experimentation. We extend it using ScenarioNet’s \cite{li_scenarionet_2023} approach, enabling the fast simulation of diverse AD datasets.
}

\contribution{We introduce V-Max, an open RL research framework for mid-to-end autonomous driving.}{Waymax \cite{gulino_waymax_2023} is an accelerated, data-driven simulator. Hardware-acceleration makes it a compelling simulator to train RL policies, however it requires re-implementing all the elements of the RL pipeline. V-Max does this work, and provides a modular pipeline, including observation functions, encoders, rewards, and algorithms. MetaDrive \cite{li_metadrive_2023} also propose tools to apply RL to the mid-to-end task, but it does not support hardware-acceleration.}

\contribution{V-Max enables the simulation of diverse datasets, it also implements various evaluation metrics and enable adversarial evaluation.}{Besides the RL training pipeline, these features aim to make V-Max a standard benchmark for mid-to-end AD. ScenarioNet \cite{li_scenarionet_2023} proposes a unified data format for AD, we adapt this approach to make datasets compatible with Waymax, enabling notably for the first time the accelerated simulation of nuPlan \cite{caesar_nuplan_2021}. We complete the evaluation metrics proposed in Waymax with the ones of the nuPlan benchmark, to provide an unified evaluation score. We include ReGentS \cite{yin_regents_2024}, a gradient-based method that generates adversarial agents, to test the robustness of driving policies.}

\contribution{Using V-Max, we conduct an experimental study of design choices in RL for AD. We end up producing highly performing RL agents. We also implement IL and rule-based baselines to show V-Max's versatility.}{We believe to be the first to perform a study of this kind, and that our findings can accelerate RL research on V-Max. We produce a Soft Actor-Critic (SAC, Haarnoja et al. (2018)) agent that solves $98\%$ of the scenarios in the non-reactive evaluation setting, demonstrating that RL can achieve strong performance in this task. However, since there is still no unified evaluation system for the task, we do not claim to be SOTA. In our final benchmark, RL dominates the other approaches, but we did not tune them as much, and did not implement the SOTA of IL. The aim of the benchmark is to show that V-Max can be used with all kind of approaches.}

\contribution{We publicly release V-Max and all the components to reproduce our experiments. We detail in the supplementary materials all the hyperparameters needed to reproduce our results.}{None}

\begin{document}

\maketitle  

\begin{abstract}
Learning-based decision-making has the potential to enable generalizable Autonomous Driving (AD) policies, reducing the engineering overhead of rule-based approaches. Imitation Learning (IL) remains the dominant paradigm, benefiting from large-scale human demonstration datasets, but it suffers from inherent limitations such as distribution shift and imitation gaps. Reinforcement Learning (RL) presents a promising alternative, yet its adoption in AD remains limited due to the lack of standardized and efficient research frameworks. To this end, we introduce V-Max, an open research framework providing all the necessary tools to facilitate RL research for AD. V-Max is built on Waymax \cite{gulino_waymax_2023}, a hardware-accelerated AD simulator designed for large-scale experimentation. We extend it using ScenarioNet’s \cite{li_scenarionet_2023} approach, enabling the fast simulation of diverse AD datasets. V-Max integrates a set of observation and reward functions, transformer-based encoders, and training pipelines. Additionally, it includes adversarial evaluation settings and an extensive set of evaluation metrics. Through a large-scale benchmark, we investigate how network architectures, observation functions, training data, and reward shaping impact RL performance. \\
Code is available at: \href{https://github.com/valeoai/v-max} {github.com/valeoai/v-max}
\end{abstract}

\section{Introduction}
\label{sec:introduction}

Reinforcement Learning (RL, \citet{sutton_reinforcement_2018}) has proven to be a powerful approach for controlling real-world systems, with milestones in dexterous robotic manipulation and industrial process control \cite{rajeswaran_learning_2018, degrave_magnetic_2022}. RL's ability to learn adaptive policies through closed-loop interaction makes it an appealing framework for Autonomous Driving (AD, \citet{kiran_deep_2022}), where decision-making agents must continuously respond to unseen scenarios and distribution shifts while maintaining high levels of robustness.

However, applying RL to real-world tasks such as AD introduces significant challenges, particularly regarding sample efficiency and training environments. As a result, RL remains underused in AD research due to practical constraints. Imitation Learning (IL, \citet{bansal_chauffeurnet_2019}) is often favored instead, as it capitalizes on vast driving datasets collected by vehicle fleets and reduces decision-making to a supervised learning task. The absence of RL-compatible environments made RL unusable in the nuPlan challenge \cite{karnchanachari_towards_2024}, which led the organizers to conclude that learning-based methods could not compete with simple rule-based baselines \cite{dauner_parting_2023}.

This gap has motivated recent efforts to improve the accessibility of RL research for AD. Notably, ScenarioNet provides an open-source framework for standardizing and replaying AD datasets in MetaDrive, an RL-compatible simulator that facilitates research on RL generalization in driving \cite{li_metadrive_2023, li_scenarionet_2023}. In parallel, \citet{gulino_waymax_2023} released Waymax, a hardware-accelerated driving simulator capable of running large-scale simulations at unprecedented speeds, making RL’s sample inefficiency less of a limiting factor for experimentation. Waymax was developed as a high-speed simulation tool, but it lacks essential benchmarking capabilities for RL research, requiring practitioners to build full training pipelines from scratch. 

In this work, we introduce V-Max, a framework that extends Waymax with all the necessary tools for RL research in autonomous driving. V-Max provides a set of observation and reward functions, multiple transformer-based encoders, and a complete training pipeline for standard RL algorithms. All these elements are implemented using the JAX framework \cite{bradbury_jax_2018}, enabling training and simulation to be performed within the same computation graph. Additionally, V-Max leverages ScenarioNet's approach to enable the accelerated simulation of diverse driving datasets, whereas Waymax was originally limited to the Waymo Open Motion Dataset (WOMD, \citet{ettinger_large_2021}). With these features, V-Max aims to standardize RL experimentation for AD, making algorithm comparisons more reproducible and accelerating progress in learning-based decision-making.

We enhance Waymax’s evaluation metrics by reimplementing nuPlan’s metrics \cite{karnchanachari_towards_2024} and introducing additional metrics, such as traffic light violations, for a more comprehensive assessment of policy performance. To further evaluate robustness, we integrate \textit{ReGentS} \cite{yin_regents_2024}, enabling evaluation against adversarial agents. We conduct a large-scale benchmark with these tools, systematically analyzing how observation functions, reward shaping, training data selection, network architectures, and learning algorithms impact performance and sample efficiency. These experiments demonstrate V-Max’s versatility, facilitating research and development on decision-making for AD. 

Our contributions are as follows:

\begin{enumerate}
    \item V-Max provides a fully integrated, JAX-based, RL training pipeline, including observation and reward functions, and transformer-based encoders inspired by motion forecasting.
    \item V-Max supports multi-dataset accelerated simulation by extending Waymax with ScenarioNet's approach.
    \item V-Max integrates comprehensive evaluation tools, including the reimplementation of nuPlan's driving quality metrics, and integration of ReGentS for adversarial evaluation.
    \item We perform a benchmark on the impact of   network architectures, observation choices, reward shaping, and training data on RL performance in AD, resulting in a policy that succesfully completes 97.86$\%$ of the scenarios in WOMD.
\end{enumerate}

\begin{figure}[t]
    \centering
    \includegraphics[width=0.95\textwidth]{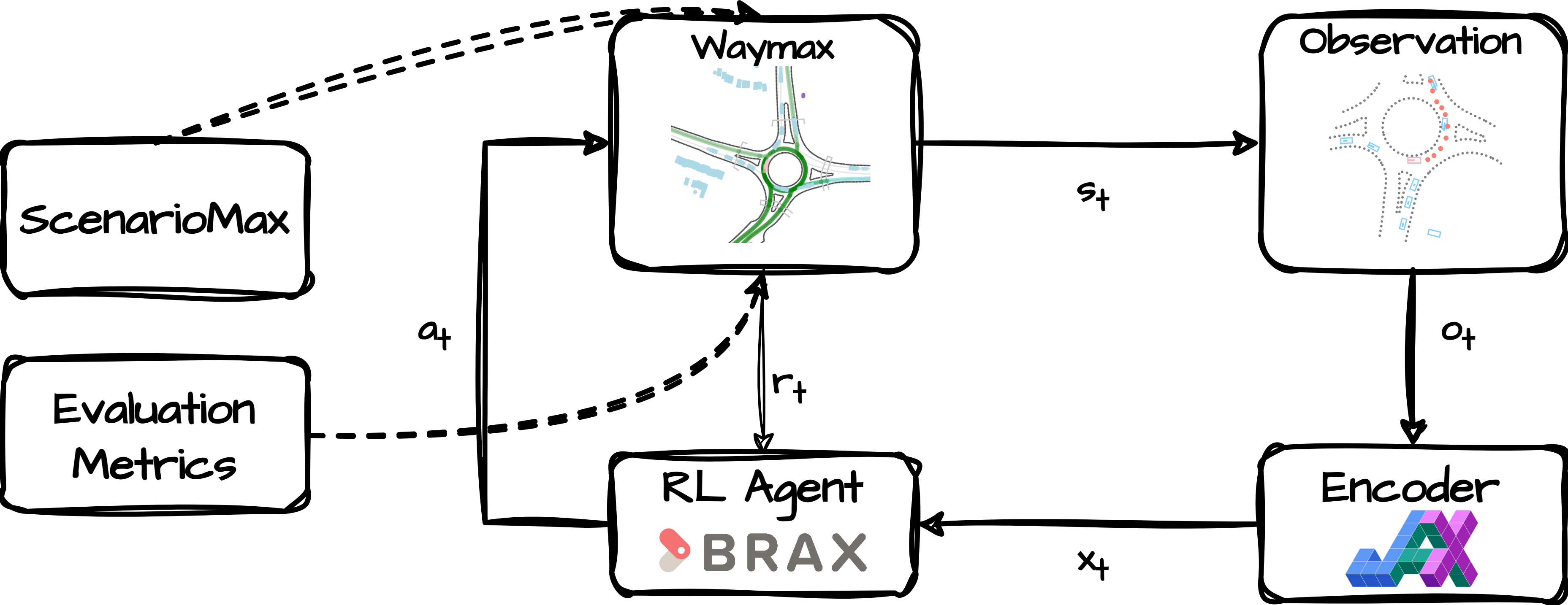}
    \caption{\textbf{Overview of the V-Max framework.} ScenarioMax standardizes multiple datasets into a Waymax-compatible format. The simulation runs in Waymax \cite{gulino_waymax_2023}, which provides the simulator state $s_t$. An observation $o_t$ is extracted and processed using a JAX-based neural encoder \cite{bradbury_jax_2018} before being fed into an RL agent implemented with Brax \cite{freeman_brax_2021}. The RL agent selects an action $a_t$ (acceleration, steering), which is executed in the simulator, receiving a reward $r_t$ based on evaluation metrics. JAX enables to run multiple instances of this process in parallel, on the same device.}
    \label{fig:vmax_framework}
\end{figure}

\section{Related Work}
\label{sec:related_work}

\subsection{Reinforcement Learning for Autonomous Driving}

There are two main formulations of the Autonomous Driving (AD) task in the literature. The first category consists of \emph{end-to-end} approaches \cite{chen_end--end_2024}, which aim to learn vehicle controls directly from raw sensor data. \citet{kendall_learning_2019} successfully applied End-to-End RL to lane-following in real-world settings, while \citet{toromanoff_end--end_2020} won the first CARLA \cite{dosovitskiy_carla_2017} challenge  using Reinforcement Learning (RL) with a supervised pretraining. These works demonstrated RL’s potential in AD, particularly as a way to overcome the limitations of Imitation Learning (IL), such as distribution shift, causal confusion and imitation gap \cite{walsman_impossibly_2022}. However, methods based solely on RL still fail to perform in the end-to-end setting, the main reason being that RL gradients are insufficient to train the large neural networks needed for perception \cite{chen_end--end_2024}. This issue is further compounded by the difficulty of creating realistic and fast simulators for the closed-loop training required for RL. Most works rely on the CARLA simulator, which allows procedurally generated scenarios to be played in the Unreal Engine \cite{dosovitskiy_carla_2017}. While generative world models such as GAIA-1 \cite{hu_gaia-1_2023} offer photorealistic closed-loop simulation, their computational cost remains a barrier to large-scale RL training.

The parallel approach is to work at mid-level and decouple the decision-making problem from the real-world perception task. In this \emph{mid-to-end} paradigm, agents process post-perception data, i.e. a structured high-level representation of the scene, and output vehicle controls. The release of large post-perception datasets like WOMD, nuScenes and Argoverse 2 \cite{caesar_nuscenes_2020, ettinger_large_2021, wilson_argoverse_2021} accelerated mid-to-end research, with a focus on the trajectory prediction sub-task. Closed-loop evaluation and training of mid-level agents was made possible with the appearance of data-driven simulators, that can replay scenarios from real-world driving while taking into account the agent's actions. Research on mid-level decision-making mainly revolves around IL and methods to improve its robustness, such as data augmentation \cite{bansal_chauffeurnet_2019}, model-based generative adversarial IL (MGAIL) \cite{bronstein_hierarchical_2022}, policy gradients \cite{scheel_urban_2022}, and curriculum learning \cite{bronstein_embedding_2023}. Notably, the \textit{nuPlan Challenge 2023} \cite{karnchanachari_towards_2024} remains the only public competition for the mid-to-end AD task, and its closed-loop challenge was won by PDM \cite{dauner_parting_2023}, a rule-based approach that significantly outperformed all the other learning-based approaches, which were all variants of imitation learning.


\citet{lu_imitation_2023} demonstrated that combining IL and RL with a simple reward signal can improve policy robustness in corner cases underrepresented in the training dataset. Similarly, \citet{grislain_igdrivsim_2024} showed that incorporating an RL objective is needed to mitigate the imitation gap, which arises from the discrepancy between the observations of human experts and those of mid-to-end AD agents (e.g. sound, turn signals). \citet{cusumano-towner_robust_2025} showed that self-play can generate highly robust policies, surpassing all prior approaches on CARLA, nuPlan, and Waymax. Their work heavily relies on a proprietary high-speed simulator, highlighting how accelerated simulation can enable large-scale RL training and significantly impact learning-based decision-making for AD. 

The successful application of RL in the simpler simulated racing task (\cite{jaritz_end--end_2018,wurman_outracing_2022}), further motivates the idea that performing simulators are key to unlock the potential of RL.

\subsection{Frameworks for mid-to-end Autonomous Driving}

V-Max is a framework built on Waymax \cite{gulino_waymax_2023} which is a data-driven, accelerated, mid-to-end AD simulator. Besides Waymax, other frameworks related to V-Max include nuPlan \cite{caesar_nuplan_2021}, Nocturne \cite{vinitsky_nocturne_2023}, MetaDrive \cite{li_metadrive_2023},  and GPUDrive \cite{kazemkhani_gpudrive_2024}.  Below, we compare them to V-Max.

\paragraph{Datasets.} All the aforementioned frameworks enable data-driven simulation, where driving scenes are instantiated by replaying real-world data. MetaDrive also integrates procedural generation, allowing to artificially instantiate driving maps and specific situations (e.g. lane merging, roundabouts). Nocturne, GPUDrive and Waymax are limited to the WOMD dataset \cite{ettinger_large_2021}, while nuPlan uses its own dataset. MetaDrive and V-Max are compatible support both nuPlan and WOMD, as well as other datasets like Argoverse 2 \cite{wilson_argoverse_2021}, thanks to the use of ScenarioNet's standardization \cite{li_scenarionet_2023}.

\paragraph{Hardware-Acceleration.} Waymax supports both acceleration on GPUs and TPUs enabling high speed simulation. If additionally the training pipeline is written using the JAX library \cite{bradbury_jax_2018}, which is the case in V-Max, then simulation and training can be performed within the same computation graph, eliminating communication bottlenecks with the host machine. GPUDrive achieves GPU-acceleration through the Madrona game engine 
 \cite{shacklett_extensible_2023}. Hardware-acceleration makes V-Max, Waymax and GPUDrive two to three orders of magnitude faster than CPU-based simulators like nuPlan, MetaDrive, and Nocturne.

\paragraph{Multi-Agent Environments.} Waymax supports environments with multiple controllable agents, a feature that V-Max uses to perform adversarial evaluation. While multi-agent RL (MARL) can technically be implemented in Waymax, V-Max is designed for traditional single-agent RL and does not include MARL-specific functionalities. In contrast, GPUDrive is explicitly designed and optimized for multi-agent learning, making it the better choice for MARL and self-play applications.

 \paragraph{Observation.} In the mid-to-end setting, simulators provide perfect perception of the scene, making the first design choice the selection of what the driving agent observes. There are two approaches to this decision. The first approach models partial observability to reduce the sim-to-real gap. Nocturne and GPUDrive use sensor-based observations that replicate camera or LiDAR properties, where vehicles can occlude one another. V-Max also implements these sensor-based observations, along with the noisy observations from IGDrivSim \cite{grislain_igdrivsim_2024}, which were designed to highlight the limitations of IL. The second approach, observation shaping, focuses on selecting an observation that maximizes policy performance while minimizing memory usage. V-Max provides tools for observation shaping and includes a comparison of different observation choices in Table \ref{tab:obs_sweep}, a topic not addressed in other frameworks.

\paragraph{Evaluation.}  MetaDrive, Nocturne, Waymax, and GPUDrive evaluate driving agents using a goal-reaching metric, which measures the percentage of scenarios where an agent successfully reaches its destination without collisions or off-road violations. nuPlan introduces a more sophisticated scoring system that also considers driving quality. V-Max integrates both the goal-reaching metric from Waymax and nuPlan’s scoring system, enabling more comprehensive evaluations and facilitating direct comparisons between agents.

\section{The V-Max Framework}
\label{sec:vmax}

\autoref{fig:vmax_framework} provides an overview of the V-Max framework, which formulates mid-to-end AD as a partially observable Markov decision process (POMDP, \citet{spaan_partially_2012}). In this section, we present the core components of V-Max and how they extend Waymax to facilitate RL's application to AD.

\subsection{Rules of the Game}

\paragraph{Simulation.} 
The simulation process leverages a \verb|simulator_state| that encapsulates data from a bird’s-eye view (BEV) representation, under the assumption that the perception problem is fully resolved. This \verb|simulator_state| includes comprehensive records of real-world scenarios, encompassing logged trajectories and high-definition (HD) maps. The primary objective of the ego vehicle is to predict control outputs, specifically acceleration and steering, to govern the vehicle’s motion from time $t$ to $t+1$. Vehicle dynamics are modeled using a continuous bicycle model, which forms the basis for motion planning and control. The simulation operates over a 9-second scenario duration, running at a frequency of 10 Hz. The initial second of each scenario is typically simulated using log-replay to establish a historical context for scene perception.

The scenario concludes when the ego vehicle violates critical safety constraints. Critical failures considered include collisions with other objects, deviations from the road, and crossing intersections under a red light. Notably, the latter constraint is not originally present in the Waymax framework and has been introduced within the V-Max framework.

\begin{figure}
    \centering
    \fbox{\includegraphics[width=0.28\linewidth]{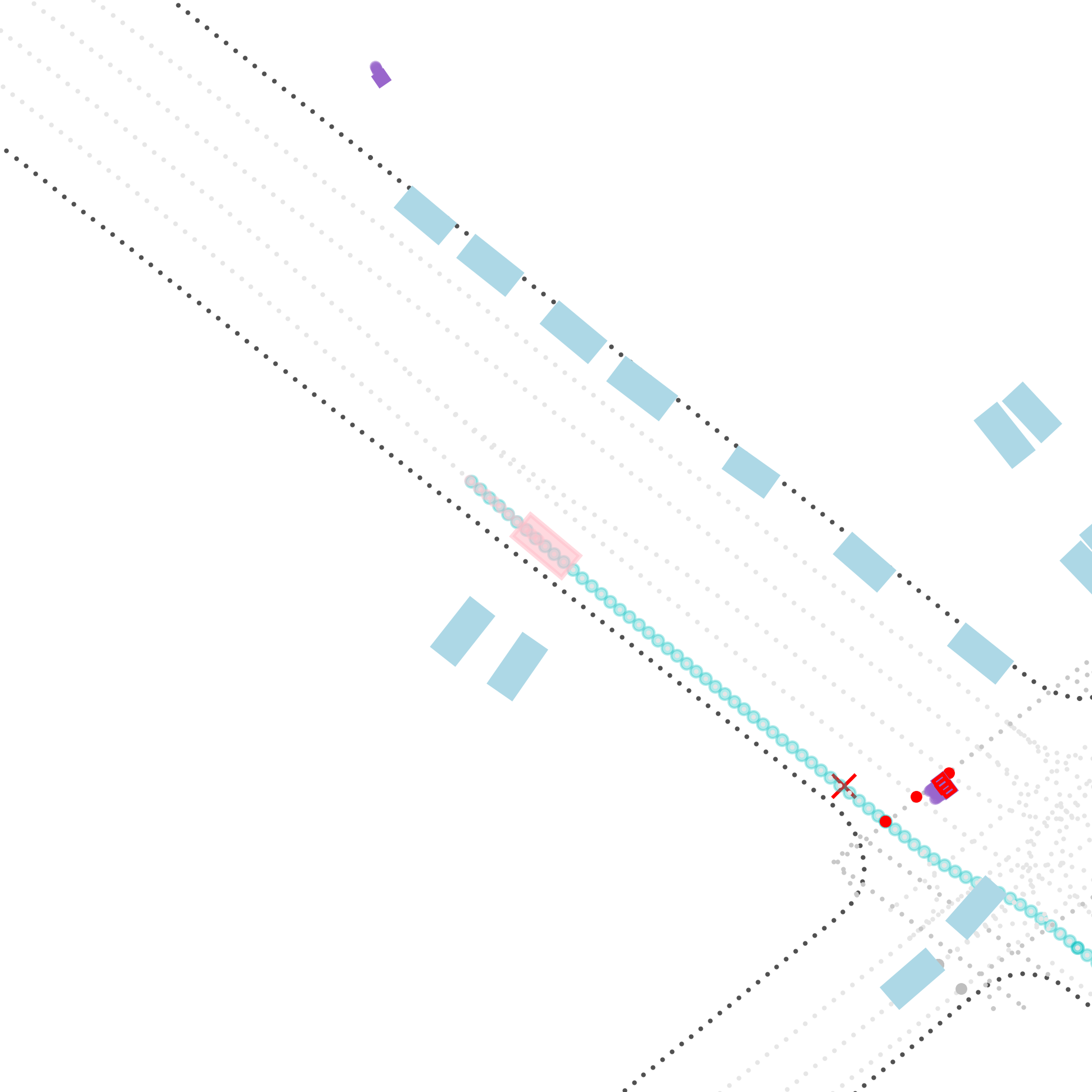}}
    \hfill
    \fbox{\includegraphics[width=0.28\linewidth]{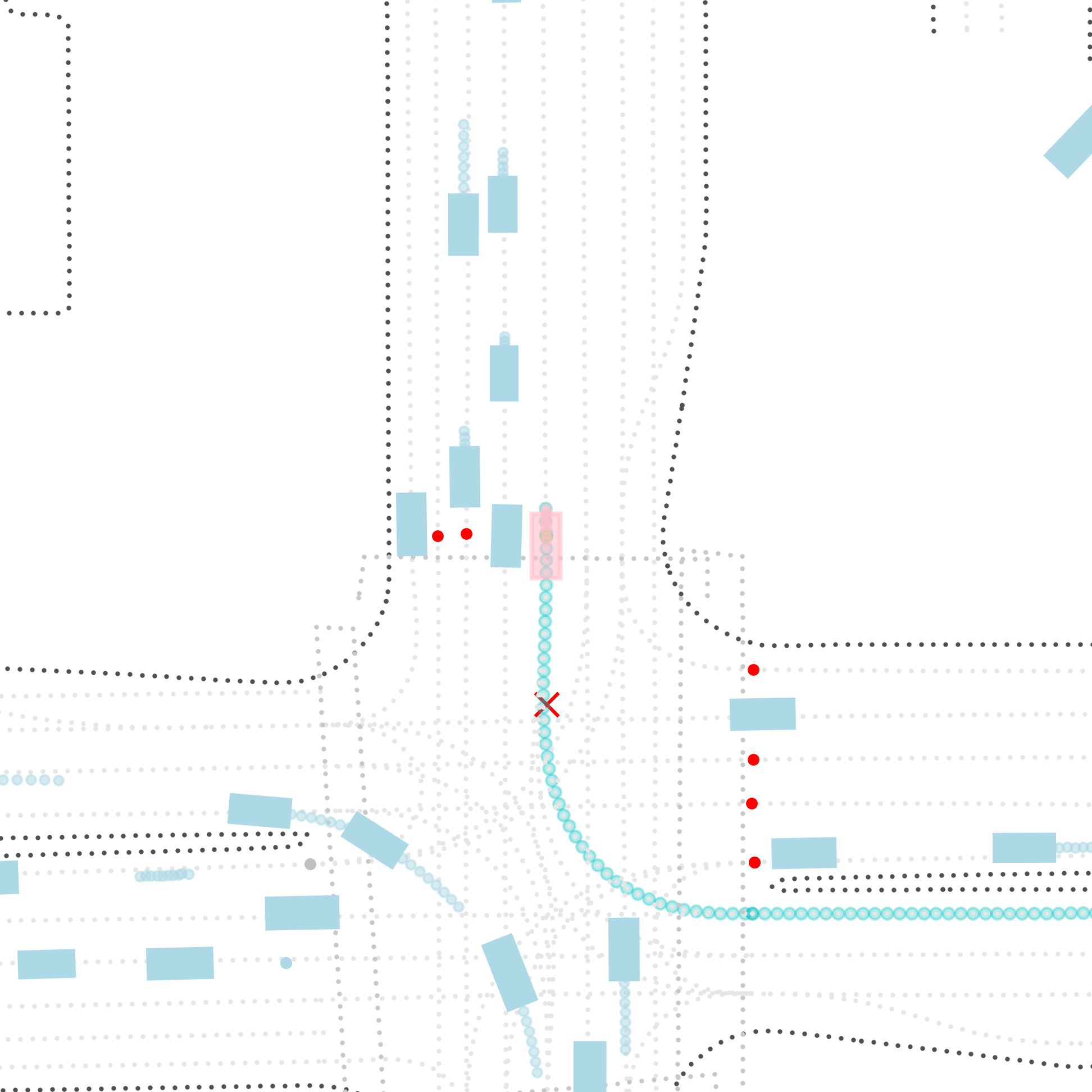}}
    \hfill
    \fbox{\includegraphics[width=0.28\linewidth]{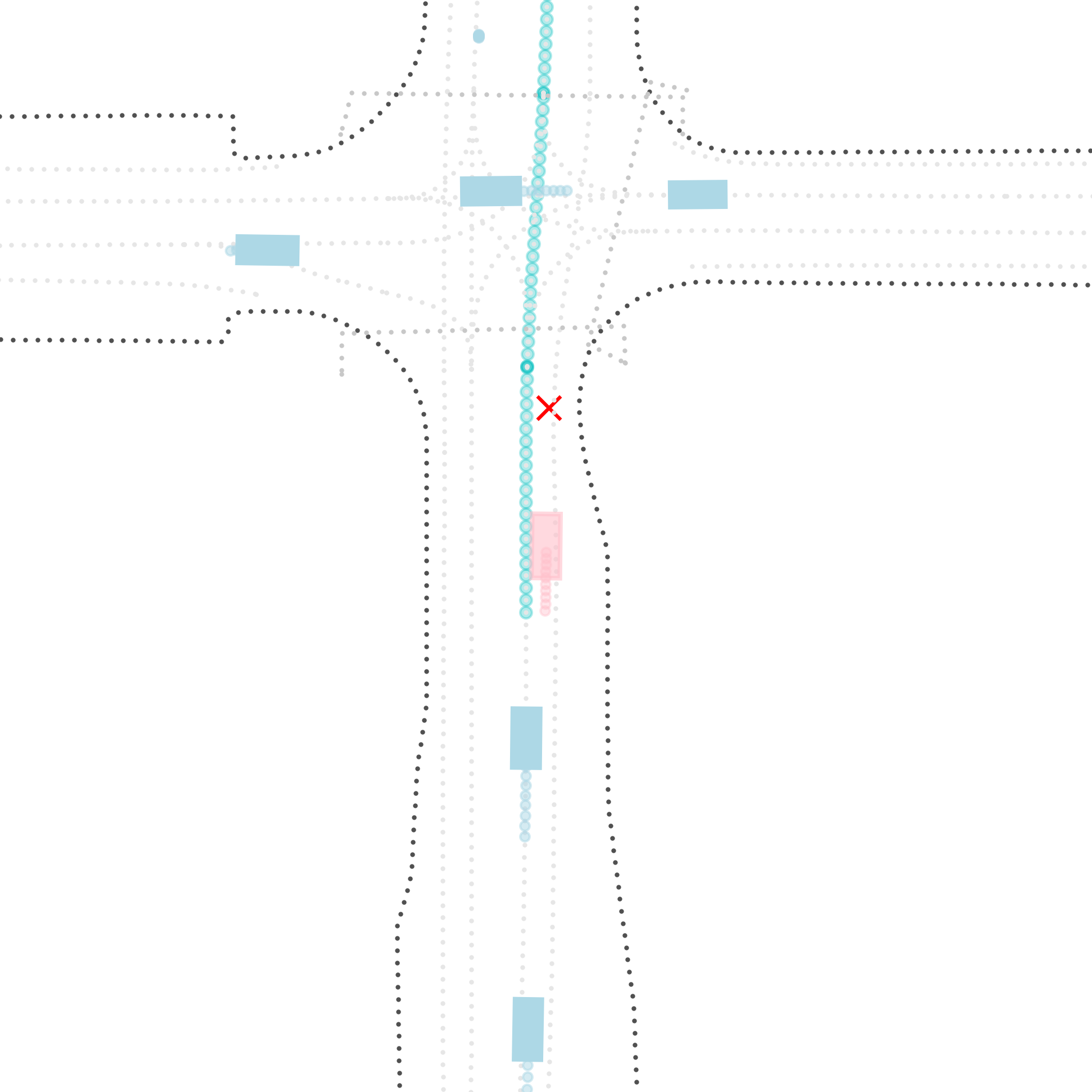}}
    \caption{\textbf{Illustration of potential limitations when using expert trajectories as learning targets.} The pink rectangle represents the ego, and the blue rectangles are the other vehicules. The blue path is one SDC path and the red cross is the last waypoint of the expert's ground truth.
    \textit{Left:} The trajectory terminates before a traffic signal, inadvertently encoding the implicit knowledge that the expert stopped at a red light. \textit{Center:} The trajectory ends in the middle of the intersection, showing that the expert stopped to let other vehicles pass, which unintentionally teaches the policy when to yield. \textit{Right:} The trajectory ends immediately prior to an intersection, which may result in the policy incidentally avoiding collisions by terminating at this location. In each scenario, we overlay the self-driving vehicle (SDC) path in blue, which provides a topologically consistent road representation without encoding such implicit behavioral biases, thus constituting a more appropriate supervisory signal for policy optimization.}
    \label{fig:goal_leakage}
\end{figure}

\paragraph{Goal.} 
V-Max does not prescribe a universal goal for the policy; instead, it allows practitioners to define the desired behavior of the ego vehicle thanks to SDC (Self-Driving Car) paths. Waymax defines SDC path as the routes given to an agent by combining the logged future trajectory of the agent with all possible future routes after the logged trajectories. At the time of writing, these paths are not publicly available in WOMD. An alternative is to rely on expert-logged trajectories only as reference paths. However, this approach is problematic because expert trajectories represent privileged information that consistently demonstrates safe behavior, as shown in Figure \ref{fig:goal_leakage}.  To overcome this limitation, V-Max enhances the simulation environment by incorporating reconstructed SDC paths. This addition enables researchers to define various tasks such as navigating to specific destinations or following predetermined routes. 

\subsection{Training RL Agents}

\paragraph{ScenarioMax.} 
One of V-Max's key contributions is ScenarioMax, an extension of ScenarioNet \cite{li_scenarionet_2023} that converts multiple open-source driving datasets into a single, compatible TfRecord format. This integration process requires several preprocessing steps to ensure data consistency and quality across different sources.

Our approach includes SDC paths reconstruction by creating drivable area definitions for the ego vehicle using road lane data. Since the original SDC paths are not publicly available, we derive them from the simulator state information. We construct paths by starting at the lane closest to the SDC's initial position, then following exit lanes. When multiple lane options exist, we create separate paths. Our method generates 10 distinct paths, selected based on their proximity to the SDC's final position. This approach captures important route options while maintaining diverse targets. Improvements could be made by adding adjacent lanes, allowing for more complex maneuvers such as safe lane changing.

While ScenarioNet proposed a scenario description format, Waymax simulator requires specific data fields to construct the \verb|simulator_state|. To address this gap, we augment the HD map data by adding directional vectors to each map point and defining proper roadgraph types. We also apply proper labeling to match the tf.Example format used by the Waymo Open Motion Dataset (WOMD).


\paragraph{Training pipeline.} V-Max uses a flexible wrapper system to encapsulate environments, drawing inspiration from the Brax \cite{freeman_brax_2021} framework's approach to parallel simulation while extending it for autonomous driving.

Notable wrappers include the AutoResetWrapper that restarts scenarios automatically when completed and the VmapWrapper that handles batched scenarios during training to accelerate policy development. We significantly modified the BraxWrapper to better integrate with our learning processes. We also added a wrapper to reconstruct one SDC path on the fly in a simulator state to support the original WOMD dataset. This wrapper is not fully recommended as it can contains errors due to the difficulty to reconstruct dynamic data in JAX jitted functions.

To support diverse learning paradigms, V-Max provides a standardized training pipeline that creates consistent agent-simulator interactions across different learning approaches (imitation learning, off-policy, and on-policy methods).  Observation and feature extraction wrappers provide a flexible mechanism for processing BEV data and state representations. The reward function module is designed for customization, allowing practitioners to define task-specific objectives and shape agent behavior through tailored incentives.

In addition to these foundational components, V-Max includes popular decision-making algorithms implementations, facilitating rapid experimentation with different policy-learning techniques. A dedicated encoder catalog further enhances the system by offering a range of neural network architectures optimized for extracting high-level representations from input features.

\begin{figure}[ht]
    \centering
    \includegraphics[width=0.9\linewidth]{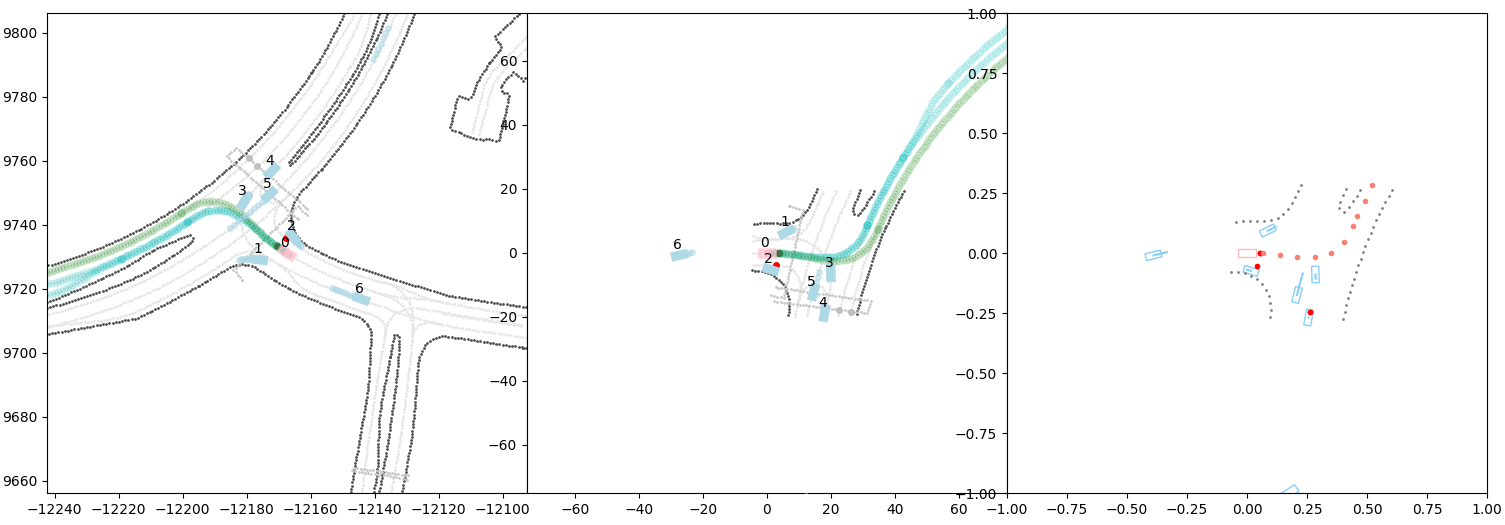}
    \caption{\textbf{Visualization of the observation process}. \textit{Left:} Scene-centered view of a scenario. SDC paths are displayed, in blue paths containing the expert trajectory, and in green showing alternative route options. \textit{Center}: Ego-centric transformation with HD map filtering within a rectangular bounding box (50 meters front, 5 meters back, 20 meters on both sides of the SDC). Optionally, noise and masking can be applied to the perception of the scene. \textit{Right:} Neural network input representation. Road boundaries are highlighted after roadgraph filtering. Only the eight closest objects to the SDC are retained, and the SDC path containing the ground truth is selected and interpolated into 10 points spaced 5 meters apart, providing a compact representation of the environment for decision making.}
    \label{fig:obs-example}
\end{figure}

\paragraph{Observation function.}
Selecting the right input features is essential for the performance of learning-based methods in autonomous driving. While Waymax provides a function to transform the simulator state to the self-driving car (SDC) view, it doesn't offer complete tools to build input features for neural networks. To solve this problem, we developed feature extractors that organize data into input features such as: (1) trajectory features showing how object motion; (2) roadgraph features describing roads and lanes; (3) traffic light features showing signal states; and (4) path target features indicating where the vehicle should go. Figure \ref{fig:obs-example} shows how the data is processed from the \verb|simulator_state| to adequate features for a neural network.

The entire feature extraction system can be customized through \verb|yaml| configurations, giving practitioners flexibility in designing observations.


\paragraph{Network architectures.} To process mid-level observations, we leverage architectures developed for motion forecasting challenges \cite{ettinger_large_2021, wilson_argoverse_2021}. These challenges focus on predicting the future trajectories of all agents in a driving scene and use the same structured scene representations as our task. Since most motion forecasting models are built on encoder-decoder architectures, their encoders can be repurposed to extract meaningful features from a mid-level driving scene, making them suitable as value and policy networks in RL algorithms.

The motion forecasting competitions are dominated by transformer-based architectures \cite{vaswani_attention_2017}. The attention mechanism is particularly useful for encoding temporal dependencies in the SDC’s past trajectory, modeling interactions between the SDC and other road users, and capturing relationships between the SDC and road features. These properties make transformers a compelling architecture for our task. While Waymax reports training results with the Wayformer  architecture \cite{nayakanti_wayformer_2023}, no official public implementation is available. We reimplement Wayformer along with other state-of-the-art encoders from the motion forecasting literature using JAX, enabling in-graph training. 

\paragraph{Reward function.} In Waymax, the reward is defined as a weighted sum of multiple components. We follow this approach and extend it by adding more reward functions based on the metrics defined in \autoref{sec:metrics}.

\subsection{Evaluation and Benchmarking}
\label{sec:metrics}

\paragraph{Metrics.} Waymax proposes the following metrics: collision rate, offroad rate, route progress ratio, and average displacement error (ADE, $\ell_2$-distance between the agent's position and the expert's position at each timestep, averaged over the trajectory). We re-implemented the metrics used in the nuPlan challenge \cite{karnchanachari_towards_2024}, which provide a more fine-grained assessment of driving quality. Notably, nuPlan distinguishes between the collisions imputable to the agent's action, and unavoidable incidents, such as rear-end collisions. Additionally, we integrate a red-light violation check, a feature absent from both Waymax and nuPlan.

\paragraph{Episode score.} To aggregate multiple metrics into a single score, we adopt the methodology from the nuPlan challenge \cite{karnchanachari_towards_2024}. Each episode is assigned a score based on a hybrid weighted average of all metric scores. The complete list of metrics and their corresponding weights are provided in the supplementary material.

\paragraph{Evaluation setups.} The main evaluation setup used in V-Max is closed-loop non-reactive, where other agents replay their logged trajectories. The advantage of this setup is that all non-controlled agents exhibit human-like behavior, as they follow real-world recorded data. However, a key limitation arises when the agent’s actions deviate from those originally taken by the expert, leading to unrealistic interactions. A common example is when an agent drives slower than the expert, causing other vehicles to collide with it from behind. This issue is partly mitigated by the short duration of scenarios (8s) and nuPlan’s distinction between at-fault and unavoidable collisions.

In order to enable reactive evaluations, where the other vehicles are controlled by a policy rather than replaying their logged trajectory, we introduce a multi-agent version of WOMD. To construct it, we first define \textit{controllable} vehicles as cars that are not offroad, and has data for the whole scenario (excluding cars that appear and disappear from the data because they were too far to be consistently detected). We then extract from the validation dataset (41872 scenarios) all scenarios with at least one controllable agent, and obtain a dataset of 31379 scenarios, with a mean of 8.72 controllable agents per scenario.

For non-ego agents, traffic lights data may be of poor quality, as the data was originally recorded from the ego vehicle, we robustify traffic lights by infering it from vehicles movement, similar to what was done in the nuPlan dataset \cite{karnchanachari_towards_2024}. \citet{yan_improving_2025} went a step further by adding missing traffic lights in intersections in the WOMD dataset, which is planned to be integrated in V-Max. 

Then, we construct SDC paths for the identified controllable agents, resulting in a new dataset, in which any policy developed on V-Max can be used to drive the controllable agents. It enables to perform closed-loop reactive evaluation with IDM policies similar to nuPlan and Waymax, excepted our IDM policies respect traffic lights, and follow the road rather than logged-trajectories.

This also enables us to try putting various policies like PDM or even evaluating how our RL policy drives when other agents use the same policy. We believe this could serve as a foundation for researchers trying to develop adversarial policies for robustness assessments.

Another evaluation setup available in V-Max applies Gaussian perturbations to the first 10 timesteps of the agent's trajectory, following the methodology of \citet{bansal_chauffeurnet_2019}. This setup assesses the policy’s ability to recover from distribution shifts, as agents in the training data are most often initialized at the center of their lane.

Finally, V-Max also integrates ReGentS \cite{yin_regents_2024}, a methodology for generating adversarial scenarios by modifying real-world driving data. In ReGentS, surrounding objects (e.g., vehicles, cyclists, and pedestrians) are optimized to create challenging situations for the agent while maintaining realistic and physically plausible interactions. The method prevents unrealistic swinging turns and unavoidable rear-end collisions, ensuring that the generated scenarios provide meaningful robustness evaluations.

\section{Case Study: RL Design Choices for Autonomous Driving}
\label{sec:case_study}

We now present the functionalities of V-Max through distinct design choices for implementing an RL agent. To isolate the impact of each component, we use the control configuration defined in \autoref{tab:default_config} and conduct experiments on observation functions, network encoders, reward shaping, and training datasets.

\begin{table}[ht]
 \caption{Control configuration used in all the experiments}
  \label{tab:default_config}
   \centering
   \small
  \begin{tabular}{cccccc}
     \toprule
    \textbf{Algorithm} &  \textbf{Encoder} & \textbf{Reward} & \textbf{Training Dataset} & \textbf{Evaluation Dataset} \\
    \midrule
    SAC  & LQ & Navigation & WOMD Training& WOMD Valid \\
    \bottomrule
  \end{tabular}
\end{table}

We replicate each run\footnote{Runs are executed on a single NVIDIA L4 GPU for 12-48 hours per run.} across 3 random seeds, and report mean and standard deviation of evaluation metrics. Accuracy denotes the percentage of episodes completed without failure conditions (collisions, off-road, or traffic signal violations). Additionally, we report collision and off-road rates, along with Waymax’s progress ratio metric, which measures how far the agent progresses relative to the expert.  Those metrics, as well as nuPlan's metrics are aggregated in the V-Max score defined in \autoref{sec:metrics_catalog}. Simulations are executed with a maximum of 64 objects per scenario to avoid memory overload.

\paragraph{Observation function.} We perform an ablation study (\autoref{tab:obs_sweep} in Appendix) to understand which BEV features are most important for a RL agent. 

We find that increasing temporal context from 1 to 5 observation steps  does not always improves V-Max scores, and adding lane center or road line features does not systematically improve performance. It suggests that road edge with waypoints information alone provides sufficient spatial context for many scenarios in the context of the simulation. The best configuration (5 steps, 16 objects, road edge features, waypoints) achieves 0.88 V-Max score compared to 0.84-0.87 for other variants. All observation functions are illustrated in \autoref{fig:observations_illustrated}.

\paragraph{Encoders.} We implemented the following encoder architectures with the Flax library \cite{heek_flax_2024}:
(1) \textbf{Latent-query} (LQ): inspired from \cite{jaegle_perceiver_2021}, (2) \textbf{Latent-query hierarchical} (LQH) \cite{bronstein_hierarchical_2022}; (3) \textbf{Motion Transformer} (MTR) \cite{shi_mtr_2024}, (4) \textbf{Wayformer} \cite{nayakanti_wayformer_2023}. For comparison, we also take an architecture that uses one multi-layer perceptron to encode each feature (road, trajectories...) separately: \textbf{MLP}. And an architecture that don't use separate encodings: \textbf{None}.

The results are reported  in Table \ref{tab:benchmark_encoders}. The Latent-query (LQ) encoder performs best across all metrics, with other transformer-based architectures (LQH, MTR, Wayformer) showing similar results. The MLP encoder performs significantly worse, confirming our intuition about the relevance of transformers-based architectures for mid-to-end AD.

\begin{table}[ht]
      \caption{Performance of encoder architectures, evaluated using the control configuration (\autoref{tab:default_config}).}
  \label{tab:benchmark_encoders}
  \centering
  \small
  \begin{tabular}{lccccc}
    \toprule
    \textbf{Encoder} & \textbf{Accuracy} $\uparrow$ & \textbf{Collision} $\downarrow$ & \textbf{Off-road} $\downarrow$ & \textbf{Progress} $\uparrow$ & \textbf{V-Max Score} $\uparrow$ \\
    \midrule
    None & 69.95±1.72 & 25.13±1.40 & 4.51±0.23 & 87.26±3.43 & 0.53±0.01 \\
    MLP & 87.54±0.48 & 9.24±0.31 & 2.92±0.24 & 104.03±2.93 & 0.68±0.01 \\
    LQ & \textbf{97.45±0.22} & \textbf{1.73±0.15} & \textbf{0.67±0.07} & 155.38±5.34 & \textbf{0.87±0.01} \\
    LQH & 96.28±0.35 & 2.52±0.12 & 1.02±0.20 & \textbf{162.23±12.26} & 0.84±0.01 \\
    MTR & 95.94±0.24 & 2.42±0.24 & 1.42±0.38 & 154.88±2.53 & 0.84±0.01 \\
    Wayformer & 96.08±0.42 & 2.70±0.41 & 0.99±0.11 & 161.94±7.20 & 0.84±0.00 \\
    \bottomrule
  \end{tabular}
\end{table}

\begin{table}[hbt]
  \caption{Performance achieved with different reward functions, evaluated using the control configuration (\autoref{tab:default_config}).}
  \label{tab:benchmark_reward}
  \centering
  \small
  \begin{tabular}{lcccccc}
    \toprule
    \textbf{Reward} & \textbf{Accuracy} $\uparrow$ & \textbf{Collision} $\downarrow$ & \textbf{Off-road} $\downarrow$ & \textbf{Progress} $\uparrow$ & \textbf{V-Max Score} $\uparrow$ & \textbf{Comfort} $\uparrow$ \\
    \midrule
    Safety & 96.73±0.57 & 2.21±0.29 & 0.90±0.34 & 78.66±3.18 & 0.67±0.04 & 57.24±1.34 \\
    Navigation & \textbf{97.45±0.22} & \textbf{1.73±0.15} & 0.67±0.07 & 155.38±5.34 & 0.87±0.01 & 68.38±3.23 \\
    Behavior & 97.25±0.31 & 1.92±0.20 & \textbf{0.64±0.05} & \textbf{139.91±4.06} & \textbf{0.88±0.02} & \textbf{70.86±2.66} \\
    \bottomrule
  \end{tabular}
\end{table}

\paragraph{Reward shaping.} We developed a hierarchical reward structure consisting of three levels: Safety (basic collision and violation penalties), Navigation (adding progress incentives), and Behavior (incorporating comfort and speed compliance). To optimize reward weights, we conducted a grid search detailed in \autoref{tab:reward_sweep} in the Appendix.

\begin{align*}
    & r_t^{\mathrm{safety}}    &&= -\mathbb{I}_{\mathrm{collision}}(t) -\mathbb{I}_{\mathrm{off-road}}(t) -\mathbb{I}_{\mathrm{red-light violation}}(t) \\
    & r_t^{\mathrm{navigation}} &&= r_t^{\mathrm{safety}} -0.2\cdot \mathbb{I}_{\mathrm{off-route}}(t) + 0.2 \cdot \mathbb{I}_{\mathrm{progress}(t)>\mathrm{progress}(t-1)} \\
    & r_t^{\mathrm{behavior}}   &&= r_t^{\mathrm{navigation}} + 0.2\cdot\mathbb{I}_{\mathrm{comfort}}(t) - 0.1\cdot\mathbb{I}_{v(t)>v_\mathrm{lim}}
\end{align*}

Results in \autoref{tab:benchmark_reward} show that the simple safety reward achieves high accuracy but results in a conservative policy, as reflected in its low progress score and overall V-Max score. Adding a navigation objective significantly improves progress while maintaining safety. The behavior reward adds an incentive to drive with respects to the speed limits and the comfort of vehicle. 

Reward shaping introduces significant optimization challenges due to competing objectives between reward components. Maximizing progress along the route while maintaining driving comfort creates a fundamental trade-off that requires precise speed control. High comfort weights can produce overly cautious policies with reduced speed, while high progression weights lead to aggressive driving that reduces safety. This demonstrates the challenge of multi-objective optimization in autonomous driving: individual reward terms can conflict with each other, which might be inefficient in order to achieve the desired driving behavior.

\paragraph{Cross-dataset experiments.} To validate ScenarioMax, we conduct a cross-dataset evaluation using agents trained on WOMD, nuPlan and combination of both. We report the results in \autoref{tab:benchmark_datasets}. Agents trained on a single dataset achieve the best performance on their respective dataset but are outperformed when evaluated on an unseen dataset, highlighting the distribution shift between WOMD and nuPlan. In contrast, the model trained on both datasets is able to achieve competitive scores on both datasets, demonstrating improved generalization.

\begin{table}[ht]
  \caption{\textbf{Preliminary study on generalization}, performance of an agent trained with the control configuration (\autoref{tab:default_config}) across nuPlan and WOMD datasets.}
  \label{tab:benchmark_datasets}
  \centering
  \small
  \begin{tabular}{cccccc}
    \toprule
    \textbf{Train Dataset} & \textbf{Accuracy} $\uparrow$ & \textbf{Collision} $\downarrow$ & \textbf{Off-road} $\downarrow$ & \textbf{Progress} $\uparrow$ & \textbf{V-Max Score} $\uparrow$ \\
    \midrule
    \multicolumn{6}{c}{\textbf{Evaluated on WOMD}} \\
    \midrule
    WOMD & \textbf{97.45±0.22} & \textbf{1.73±0.15} & \textbf{0.67±0.07} & 155.38±5.34 & \textbf{0.87±0.01} \\
    nuPlan & 76.81±1.53 & 7.30±1.25 & 1.30±0.28 & 163.77±8.15 & 0.67±0.01 \\
    WOMD + nuPlan & 96.24±0.68 & 2.49±0.57 & 0.98±0.10 & \textbf{172.29±10.28} & 0.85±0.02 \\
    \midrule
    \multicolumn{6}{c}{\textbf{Evaluated on nuPlan}} \\
    \midrule
    WOMD & 87.73±0.64 & \textbf{1.89±0.20} & 2.66±0.15 & 308.64±4.04 & 0.76±0.01 \\
    nuPlan & 95.33±0.42 & 2.27±0.23 & 1.86±0.17 & \textbf{319.84±14.02} & \textbf{0.82±0.00} \\
    WOMD + nuPlan & \textbf{95.38±0.71} & 2.04±0.40 & \textbf{1.98±0.21} & 315.95±15.77 & \textbf{0.82±0.01} \\
    \midrule
    \multicolumn{6}{c}{\textbf{Evaluated on WOMD + nuPlan}} \\
    \midrule
    WOMD & 94.21±0.05 & \textbf{1.80±0.13} & 1.42±0.07 & 211.43±3.18 & 0.83±0.01 \\
    nuPlan & 82.76±1.16 & 5.69±0.92 & 1.47±0.22 & 213.79±9.95 & 0.72±0.01 \\
    WOMD + nuPlan & \textbf{95.97±0.68} & 2.35±0.51 & \textbf{1.29±0.13} & \textbf{218.33±11.96} & \textbf{0.84±0.01} \\
    \bottomrule
  \end{tabular}
\end{table}

\section{Benchmark}

We now present a benchmark of various approaches reimplemented in V-Max and report the best performance achieved by each method. Each method is executed across three random seeds, and the policy with the highest V-Max score is selected. The hyperparameters used for training are detailed in \autoref{sec:hyperparameters}.

\subsection{Standard evaluation}
\label{bench:planning}

We first compare approaches in the closed-loop non-reactive evaluation setup. PPO \cite{schulman_proximal_2017} and SAC \cite{haarnoja_soft_2018} represent RL approaches, IDM \cite{treiber_congested_2000} and PDM \cite{dauner_parting_2023} represent rule-based approaches, more details on our implementation of PDM can be found in \autoref{sec:PDM}. Note that we used \textit{PDM-closed}, as it has the best scores in closed-loop evaluation and is purely rule-based. To represent IL, we implemented BC with a Wayformer architecture, similar to the one used in Waymax's benchmark. Finally, we also reimplemented BC-SAC from \cite{lu_imitation_2023}, trained with the Safety reward.

Notably, PPO and BC train significantly faster than SAC, so we allow them to run for a higher number of timesteps to ensure fair comparisons at equivalent training durations. All the trainings and evaluations are made using WOMD.

\begin{table*}[ht]
  \caption{\textbf{V-Max benchmark} - evaluations on WOMD Valid, in the closed-loop non-reactive setting. A complete definition of metrics is provided in \autoref{sec:metrics_catalog}.}
  \label{tab:big_benchmark}
  \centering
  \small
  \begin{tabular}{lc|cc|cccc}
    \toprule
    \textbf{Metric} & \textbf{Expert} & \textbf{IDM} & \textbf{PDM} & \textbf{SAC} & \textbf{BC+SAC} & \textbf{PPO} & \textbf{BC} \\
    \midrule
    Collision rate $\downarrow$& 0.49 & 2.78 & 0.96 & \textbf{0.89} & 1.25 & 7.81 & 13.14 \\
    Offroad $\downarrow$ & 0.53 & 2.47 & 1.05 & \textbf{0.65} & 0.55 & 1.14 & 6.92 \\
    Red-light violation $\downarrow$ & 0.65 & 0.26 & 0.21 & 0.21 & \textbf{0.20} & 0.33 & 0.55 \\
    Progress $\uparrow$ & 98.94 & 144.44 & 140.58 & 170.85 & 90.98 & \textbf{189.52} & 86.87 \\
    At-fault collisions $\downarrow$ & 0.49 & 2.78 & 0.96 & \textbf{0.89} & 1.25 & 5.57 & 4.49 \\
    Making progress $\uparrow$ & 99.45 & 97.11 & 97.78 & \textbf{99.02} & 90.49 & 97.58 & 87.97 \\
    Driving direction compliance $\uparrow$ & 97.82 & \textbf{99.66} & 99.52 & 98.11 & 90.95 & 99.26 & 97.47 \\
    TTC within bound $\uparrow$ & 99.16 & 92.86 & 95.90 & 96.12 & \textbf{96.96} & 83.63 & 94.79 \\
    Progress along route ratio $\uparrow$ & 99.18 & 92.65 & 92.17 & \textbf{95.13} & 75.60 & 94.61 & 78.64 \\
    Speed limit compliance $\uparrow$ & 99.91 & 99.67 & \textbf{99.70} & 97.69 & 99.59 & 98.35 & 98.55 \\
    Multiple lanes score $\uparrow$ & 90.96 & \textbf{99.25} & 96.28 & 90.62 & 72.65 & 94.00 & 89.68 \\
    Comfort  $\uparrow$ & 96.92 & 69.75 & 58.05 & 67.84 & 68.95 & 18.69 & \textbf{93.32} \\
    \midrule
    Accuracy $\uparrow$ & 98.26 & 89.96 & 93.60 & \textbf{97.86} & 97.40 & 90.75 & 79.42 \\
    Accuracy only at-fault $\uparrow$ & 98.33 & 94.51 & 97.78 & \textbf{98.25} & 98.02 & 88.07 & 92.99 \\
    V-Max score $\uparrow$ & 0.945 & 0.876 & 0.887 & \textbf{0.892} & 0.713 & 0.784 & 0.721 \\
    \bottomrule
  \end{tabular}
\end{table*}

Results of the benchmark are reported in \autoref{tab:big_benchmark}. The expert serves as an upper bound for accuracy, as failures only occur due to data errors. The poor performance of BC and PPO is likely due to limited hyperparameter tuning, as we concentrated our efforts on optimizing the SAC baseline.

We see that V-Max enabled us to train a highly performing driving policy using a standard RL algorithm, achieving an accuracy of $97.86\%$ and strong performance across all sub-metrics of the V-Max score, using a single NVIDIA L4 GPU for 48 hours.

Although a RL planner can achieve expert-like accuracy in simulation, we see that our agents still fail to optimize for the comfort metric and the overall V-Max score, which could be investigate in future works in order to push a RL planner im the real world.

\subsection{Reactive Evaluations}
\label{bench:evaluations}

Table \ref{tab:reactive} presents our first experiment on using our multi-agent dataset for reactive evaluations. We see that safer models, like PDM and our SAC policy, allow to reduce collision rates for all policies. We specifically see how we managed to mitigate the weakness of log-replay by looking at the "Other collisions" raw, that account for collisions not imputable to the ego (rear-bumper collisions mainly).

\begin{table}[htbp]
\centering
\caption{\textbf{Reactive evaluations} – Results on the first 1000 scenarios of our multi-agent dataset. Expert corresponds to the non-reactive setting.}
\label{tab:cross_algorithm_results_reordered_metrics}
\resizebox{\textwidth}{!}{%
\begin{tabular}{@{}l|cccc|cccc|cccc@{}}
\toprule
\multirow{2}{*}{Metric} &
  \multicolumn{4}{c|}{\textbf{IDM vs}} &
  \multicolumn{4}{c|}{\textbf{PDM vs}} &
  \multicolumn{4}{c}{\textbf{SAC vs}} \\
\cmidrule(lr){2-5} \cmidrule(lr){6-9} \cmidrule(lr){10-13}
& \textbf{Expert} & \textbf{IDM} & \textbf{PDM} & \textbf{SAC} &
  \textbf{Expert} & \textbf{IDM} & \textbf{PDM} & \textbf{SAC} &
  \textbf{Expert} & \textbf{IDM} & \textbf{PDM} & \textbf{SAC} \\
\midrule
Accuracy $\uparrow$
  & 89.7 & 91.0 & 96.0 & 95.5
  & 94.3 & 95.1 & 98.4 & 97.8
  & 97.7 & 95.9 & \textbf{98.5} & 98.4 \\

At-Fault Collision $\downarrow$
  & 2.4 & 3.4 & 1.9 & 1.8
  & 0.2 & 0.4 & \textbf{0.1} & 0.3
  & 0.7 & 1.1 & 0.6 & 0.4 \\

Other collisions $\downarrow$
  & 6.10 & 4.10 & 0.20 & 1.00
  & 4.70 & 3.20 & 0.30 & 0.90
  & 0.60 & 2.00 & \textbf{0.10} & 0.50 \\
\bottomrule
\end{tabular}%
}
\label{tab:reactive}
\end{table}

\subsection{Adversarial Evaluation}

Table \ref{tab:regents} showcases how ReGentS can be used to perform adversarial evaluations. The ReGentS agents are optimized to have a trajectory that aggressively cuts-in the ego. We see that the RL agent, thanks to its closed-loop training, is far more robust than the other counterparts. 

However, as it could be expected, most of the collisions created by ReGentS are classified as not imputable to the ego, but they should be taken into account in this setting, as such aggressive behaviors can occur in real life.

\begin{table}[htbp]
\centering
\small
\caption{\textbf{Adversarial evaluations} – Results on the first 100 valid scenario for ReGentS.}
\label{tab:model_setting_comparison}
\begin{tabular}{@{}l|cc|cc|cc@{}}
\toprule
\multirow{2}{*}{Metric} &
  \multicolumn{2}{c|}{\textbf{IDM}} &
  \multicolumn{2}{c|}{\textbf{PDM}} &
  \multicolumn{2}{c}{\textbf{SAC}} \\
\cmidrule(lr){2-3} \cmidrule(lr){4-5} \cmidrule(lr){6-7}
& Non Reactive & ReGentS
& Non Reactive & ReGentS
& Non Reactive & ReGentS \\
\midrule
Accuracy $\uparrow$
  & 92 & 43
  & 96 & 53
  & 98 & \textbf{76} \\

At‑Fault Collision $\downarrow$
  & 1 & \textbf{5}
  & 0 & 11
  & 2 & 11 \\

Other collisions $\downarrow$
  & 5 & 51
  & 4 & 33
  & 0 & \textbf{10} \\
\bottomrule
\end{tabular}
\label{tab:regents}
\end{table}

\section{Conclusion}
\label{sec:conclusion}

In this work, we introduced V-Max, a framework designed to facilitate Reinforcement Learning research for mid-to-end Autonomous Driving (AD). Built on Waymax, V-Max extends its capabilities with a JAX-based RL training pipeline, multi-dataset accelerated simulation, and comprehensive evaluation tools.  Using these tools, we trained high-performing SAC agents, showing how V-Max can help advance RL research for AD. To further support progress in this area, we ensure full reproducibility by publishing our framework and benchmarks.

While V-Max provides a foundation for AD research, rigorously evaluating driving policies remains an open challenge. Current evaluation protocols (in V-Max and the frameworks discussed in \autoref{sec:related_work}) average scenario metrics across the entire validation dataset. However, driving difficulty follows a long-tail distribution \cite{makansi_exposing_2021, bronstein_embedding_2023}, where most scenarios are easily solvable while a small subset presents significant challenges. Developing benchmarks that explicitly account for this distribution would enable a more rigorous assessment of policy robustness.

Additionally, further research on adversarial scenario generation, could enable deeper robustness assessment of driving policies. ReGentS is a good starting point, diffuser-based methods is an alternative approach \cite{pronovost_scenario_2023}.  Similarly, the development of more realistic simulation agents, as explored in the \textit{Waymo Open Sim Agents Challenge} (WOSAC, \cite{montali_waymo_2023}) could improve realism of closed-loop simulators, reducing the reliance on non-reactive evaluation.



\bibliography{main}
\bibliographystyle{rlj}

\beginSupplementaryMaterials

\appendix

\section{Perfomances}

We run a performance comparison between Waymax and V-Max, measuring steps per second (SPS) when replaying expert trajectories. The benchmark task involves converting expert steering and acceleration commands through Waymax's InvertibleBicycle model while incrementally adding V-Max's observation extraction and metrics computation within the environment step. All measurements were conducted on a single Nvidia L4 GPU with 16 CPUs and 64GB RAM. For fair comparison, Waymax's default observation function was overridden to return a dummy array, as the full state observation typically reduces performance by 30-40\%. 

\begin{table}[h]
\centering
\caption{SPS comparaison between Waymax and V-Max features.}
\begin{tabular}{lcc}
\toprule
& \textbf{Waymax} & \textbf{V-Max} \\
\midrule
\multicolumn{3}{l}{\textit{Single scenario, 64 objects}} \\
\midrule
Env & 623.08 & 615.63 \\
Env + Obs & N/A & 543.55 \\
Env + Metrics & N/A & 367.46 \\
Env + Metrics + Obs & N/A & 342.72 \\
\midrule
\multicolumn{3}{l}{\textit{Batch of 32 scenarios, 64 objects}} \\
\midrule
Env & 10109.73 & 9942.12 \\
Env + Obs & N/A & 8210.96 \\
Env + Metrics & N/A & 5387.38 \\
Env + Metrics + Obs & N/A & 4609.01 \\
\bottomrule

\end{tabular}
\label{tab:sps}
\end{table}

\section{Metrics}
\label{sec:metrics_catalog}

\textbf{Waymax Metrics}

\begin{itemize}
    \item Offroad: Binary flag indicating whether the SDC left the drivable area at any point in the scenario.
    \item Collision (overlap): Binary flag indicating whether the SDC collided with another object at any point in the scenario.
    \item Wrongway: Waymax-specific metric based on SDC paths, indicating whether the SDC is more than 3.5 meters away from the closest SDC path.
    \item Offroute: Similar to wrongway but with respect to on-route paths, which are the SDC paths that contain the expert’s logged trajectory.
    \item \verb|sdc_progress|: computes how much the SDC progressed along on-route paths, and divides it by the distance the expert did cover on those paths. Can be greater than 1.
\end{itemize}

For example, if the SDC takes a right turn at an intersection while the expert proceeded straight, the SDC will be considered offroute but not wrongway.

\textbf{nuPlan Metrics}
\begin{itemize}
    \item Progress along route: same definition as Waymax, but capped to 1. 
    \item Making progress: binary flag indicating if progress along route is superior to 20$\%$. 
    \item At-fault collision: Binary flag following nuPlan's criteria for assigning collision responsibility:
    \begin{itemize}
        \item Collisions with stopped vehicles are always at-fault.
        \item If the SDC is stopped, it is never at-fault.
        \item If the SDC is occupying multiple lanes, it is at-fault.
        \item Rear-bumper collisions are not at-fault, while front-bumper collisions are at-fault.
    \end{itemize}
    \item TTC within bound: indicates if the time-to-colllision (ttc) with ahead vehicles remain superior to $0.95$s.
    \item Speed limit compliance: defined by nuPlan as:
    \begin{equation}
        \text{nuplan\_speed\_compliance} = \max \left(0.0, 1.0 - \frac{\sum_t \text{speed\_violation}_t \cdot \Delta t}{\max(\text{threshold}, 1e-3) \cdot T} \right),
    \end{equation}
    where $\text{speed\_violation}_t$ is the magnitude of overspeeding at timestep $t$, $\Delta t$ is the time step duration, and $T$ is the total scenario duration.
    \item Driving direction compliance: Based on distance traveled into oncoming traffic. We check if the vehicle is effectively driving into oncoming traffic lanes using the road information, rather than SDC paths.
    \begin{itemize}
        \item Score = 1.0 if wrong-way distance $\leq 2.0$m.
        \item Score = 0.5 if wrong-way distance is between 2.0m and 6.0m.
        \item Score = 0 if wrong-way distance $> 6.0$m.
    \end{itemize}
 \item  Comfort: binary indicating if the trajectory is comfortable based on jerk, acceleration and yaw rates.
\end{itemize}

\textbf{V-Max Metrics}
\begin{itemize}
    \item Red-light violation: Binary flag indicating whether the SDC crossed an intersection while the traffic light was red.
    \item Time spent on multiple lanes: Evaluated based on roadgraph information rather than SDC paths. We added this metric to encourage agent to remain on one lane, to set the thresholds, we looked at the expert trajectories.
    \begin{itemize}
        \item Score = 1.0 if time spent on multiple lanes $\leq 3.4$s.
        \item Score = 0.5 if time spent on multiple lanes is between 3.4s and 5.7s.
        \item Score = 0 if time spent on multiple lanes $> 5.7$s.
    \end{itemize}
\end{itemize}

\begin{table}[ht]
  \caption{Metrics and their weights in nuPlan aggregate score and V-Max aggregate score. $^\dagger$ indicates metrics that appear only in the V-Max aggregate score.}
  \label{tab:metrics_weights}
  \centering
  \small
  \begin{tabular}{lcc}
    \toprule
    \textbf{Metric  name} & \textbf{Multiplier weight} & \textbf{Average weight} \\
    \midrule
    No at-fault collisions & $\{0, 1\}$ & - \\
    Offroad & $\{0, 1\}$ & - \\
    Red-light violation$^\dagger$ & $\{0, 1\}$ & - \\
    Making progress & $\{0, 1\}$ & - \\
    Driving direction compliance & $\{0, 0.5, 1\}$ & - \\
    TTC within bound & - & 5 \\
    Progress along route ratio & - & 5 \\
    Speed limit compliance & - & 4 \\
    Multiple lanes score$^\dagger$ & - & 3 \\
    Comfort & - & 2 \\
    \bottomrule
  \end{tabular}
\end{table}

\section{Reimplementing PDM in V-Max}
\label{sec:PDM}

We first reimplement IDM \cite{treiber_congested_2000}, using the path target as a reference path, and considering the first object or red light along this path as the leading agent. This baseline is already quite strong. Then PDM-Closed \cite{dauner_parting_2023}, generates multiple trajectories by using PDM with the target path and two offset paths ($\pm 1.0$ meter) and five desired speeds, namely $\{20\%, 40\%, 60\%, 80\%, 100\%\}$ of the speed limit.

External objects are forecasted using a constant-speed and constant-heading assumption, and each candidate trajectory is scored based on this forecast. The first action of the highest-scoring trajectory is then executed, and the process repeats at each timestep, which might explain the low comfort metric observed in \autoref{tab:big_benchmark}.

Following \citet{dauner_parting_2023}, we introduce an emergency braking maneuver when the estimated time-to-collision (TTC) falls below 2.0s. While this improves PDM’s evaluation scores, it also makes the behavior less realistic.

\section{Observation functions}

We present and details the observation functions used in the experiments. We tested different observations represented in \autoref{tab:obs_sweep}. For object features, the module extracts waypoints, velocity, yaw, size, and validity information from the $n$ closest objects to the ego vehicle. Roadgraph features include waypoints, direction, speed limits, and validity data, sampled at 2-meter intervals within a maximum distance of 50 meters. The module selects the top 200 roadgraph elements within a bounding box extending 50 meters forward, 5 meters back, and 20 meters to each side. Traffic light features capture waypoints, state, and validity information for the 5 closest traffic lights. Finally, path target features consist of waypoints for the target trajectory, sampled from the SDC paths.

\begin{figure}[ht]
    \centering    
    \begin{subfigure}[t]{0.35\textwidth}
        \centering
        \fbox{\includegraphics[width=0.8\textwidth]{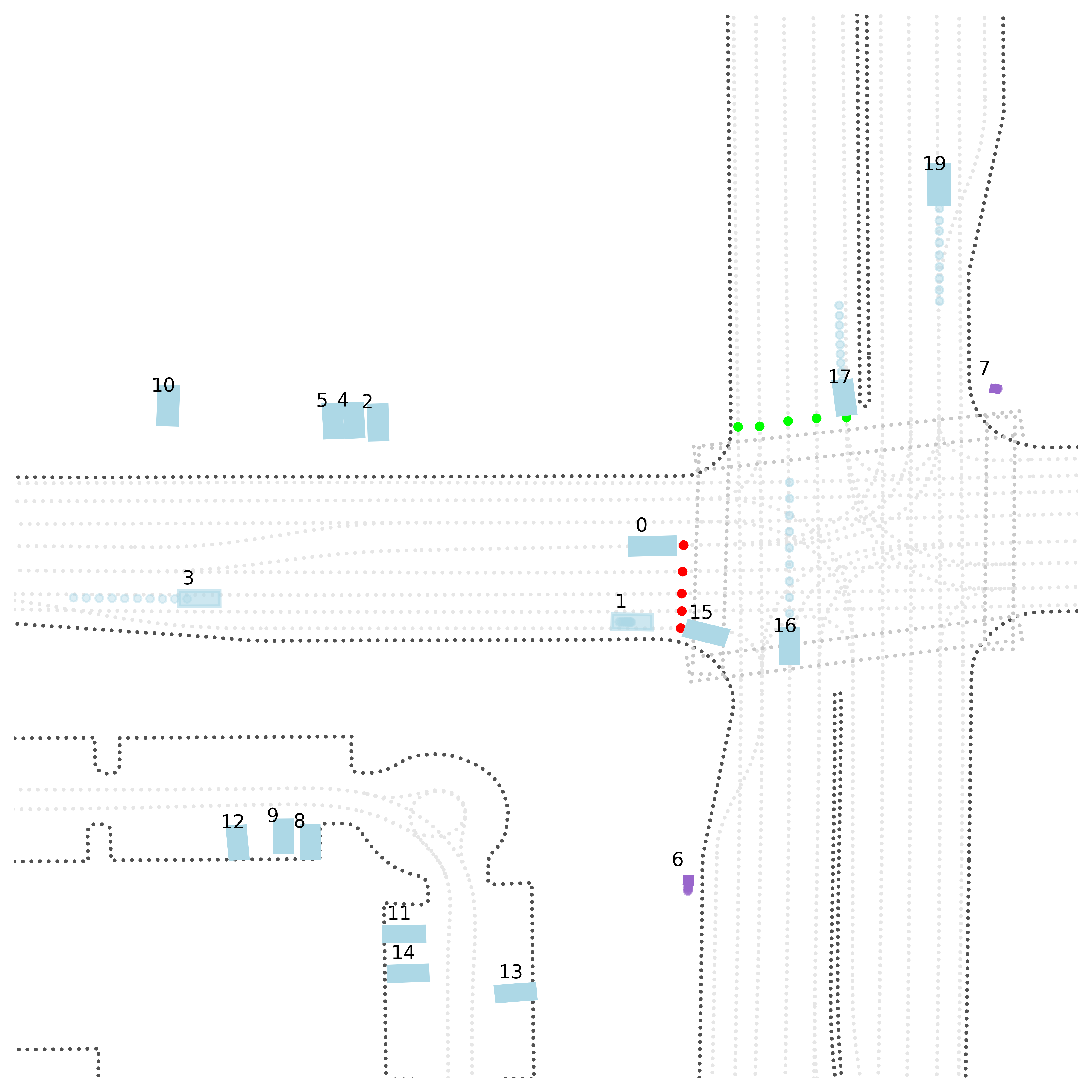}}
        \caption{Simulator state}
        \label{fig:state_obs}
    \end{subfigure}
    
    \vspace{0.1cm}
    
    \begin{minipage}[t]{0.48\textwidth}
        \centering
        \small\textbf{10x5 Waypoints}
        
        \begin{subfigure}[t]{\textwidth}
            \centering
            \fbox{\includegraphics[width=0.6\textwidth]{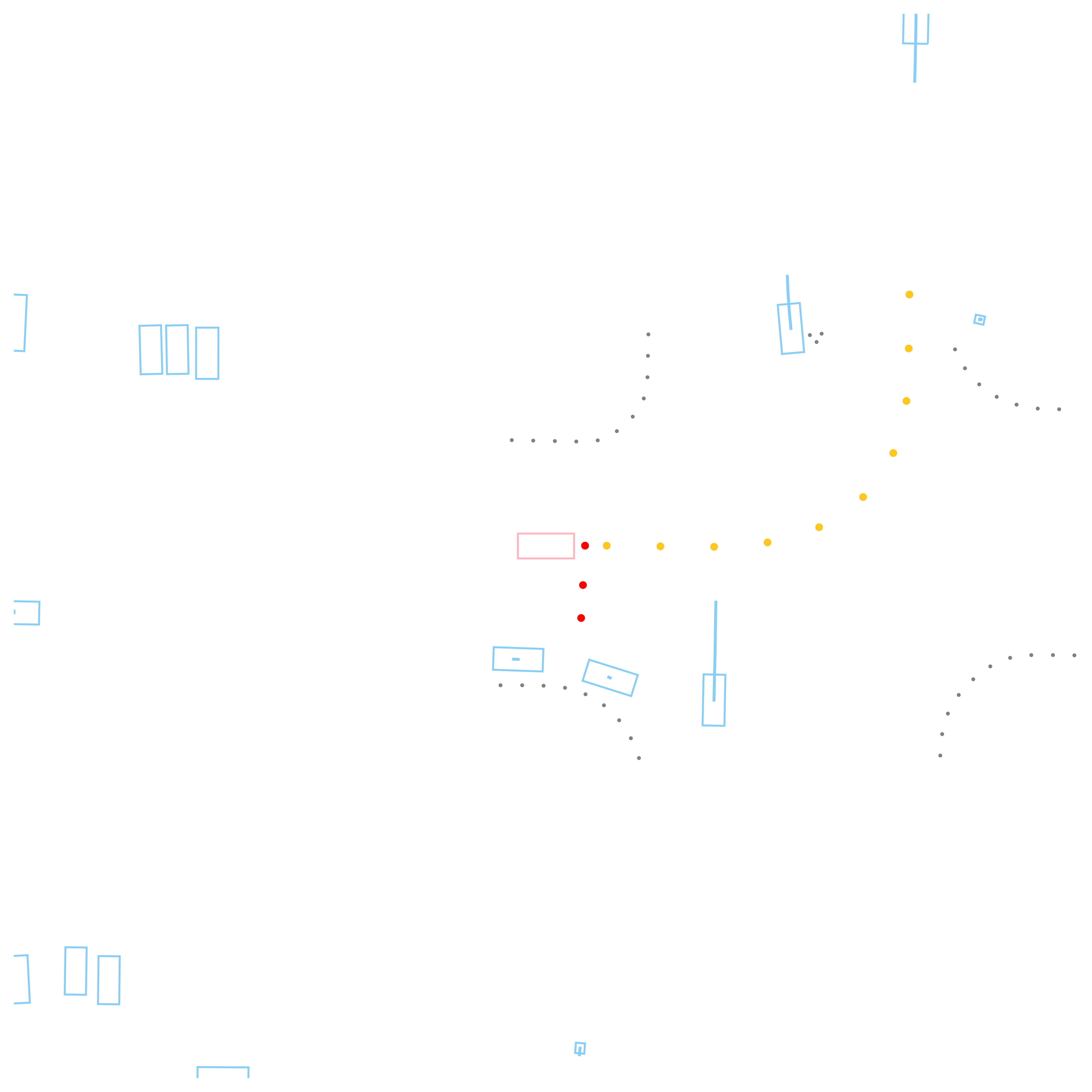}}
            \caption{road edge}
            \label{fig:segment_obs}
        \end{subfigure}
        
        \begin{subfigure}[t]{\textwidth}
            \centering
            \fbox{\includegraphics[width=0.6\textwidth]{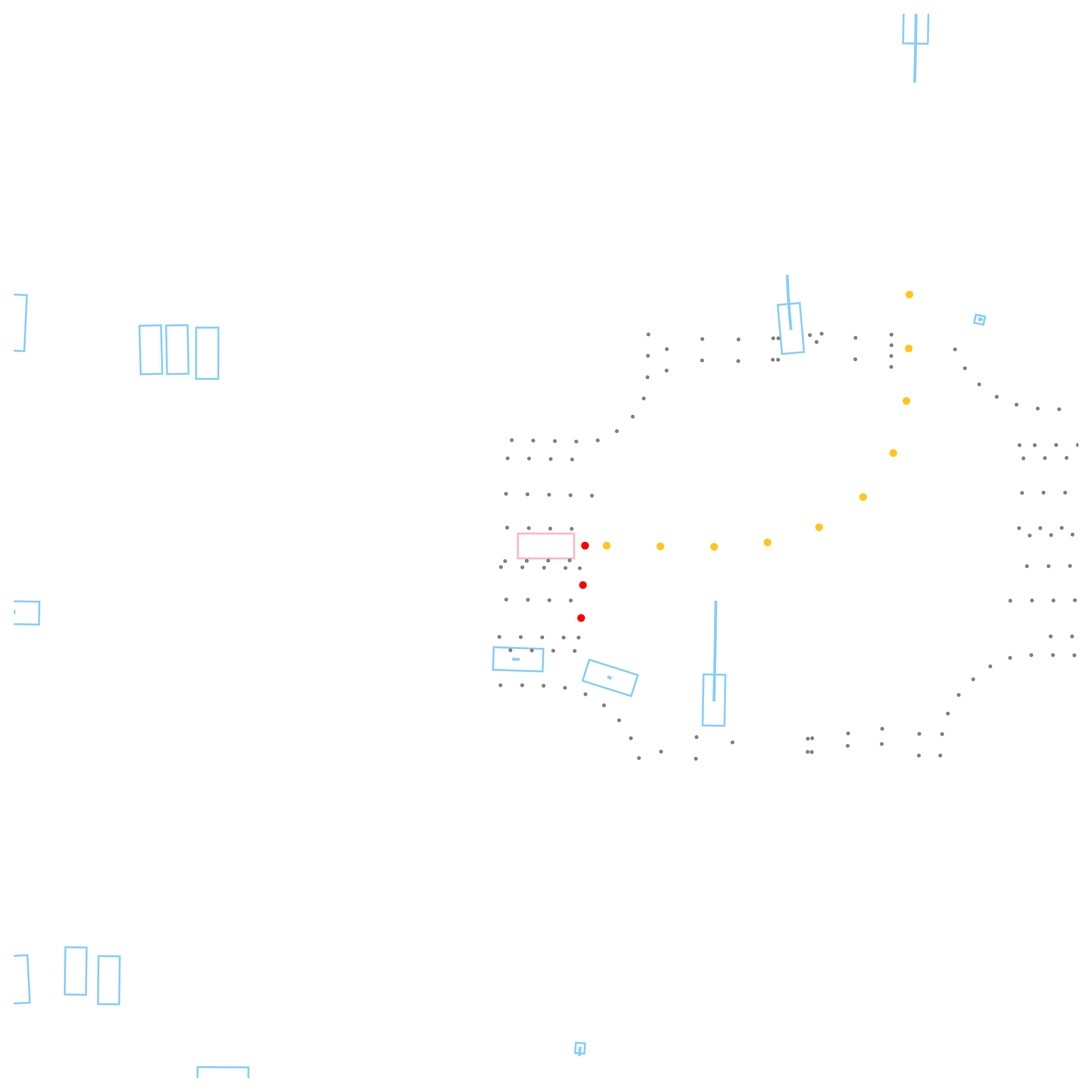}}
            \caption{road edge, road line}
            \label{fig:road_obs}
        \end{subfigure}
        
        \begin{subfigure}[t]{\textwidth}
            \centering
            \fbox{\includegraphics[width=0.6\textwidth]{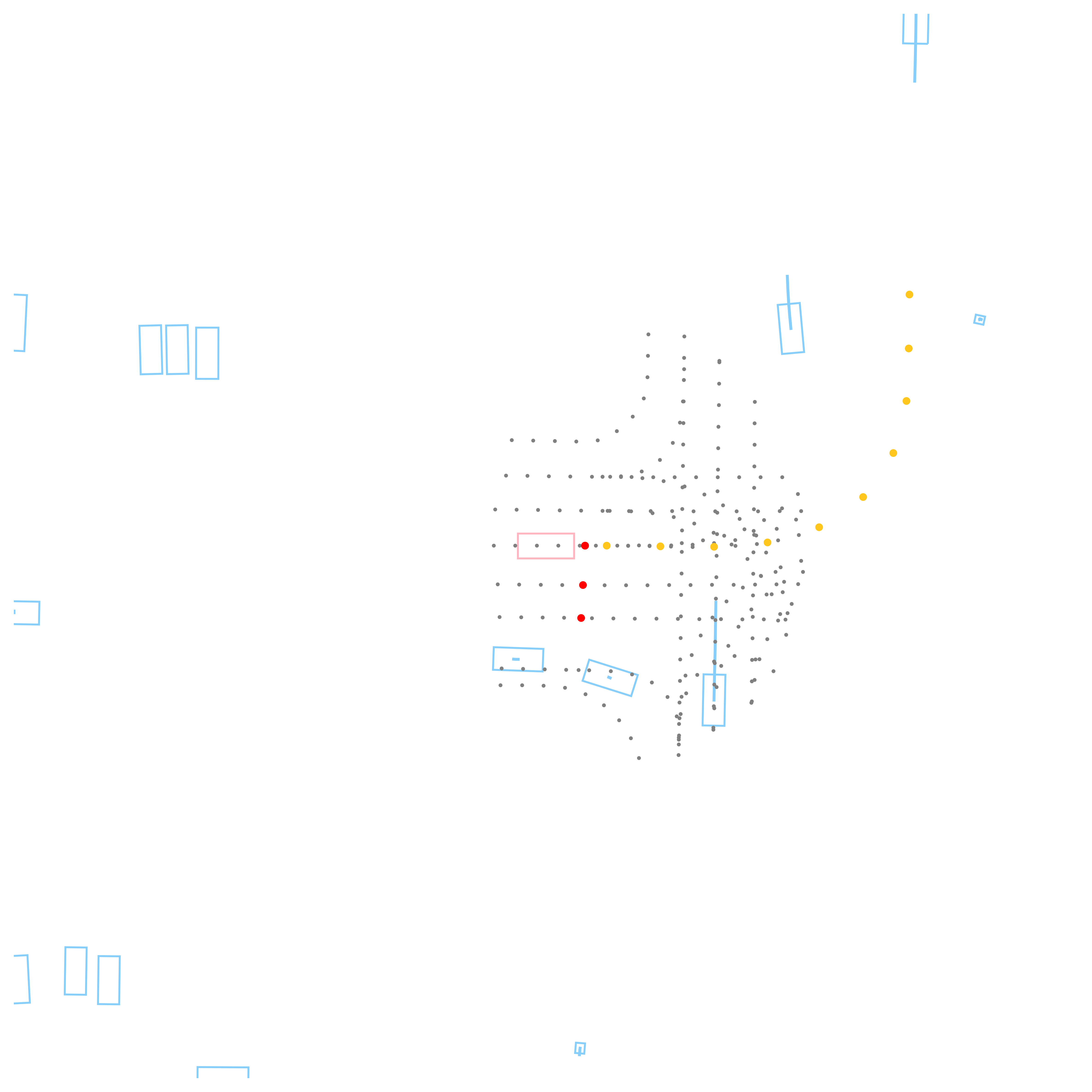}}
            \caption{road edge, lane center}
            \label{fig:road_obs_waypoints}
        \end{subfigure}
    \end{minipage}
    \hfill
    \begin{minipage}[t]{0.48\textwidth}
        \centering
        \small\textbf{1x30 Goal Point}
        
        \begin{subfigure}[t]{\textwidth}
            \centering
            \fbox{\includegraphics[width=0.6\textwidth]{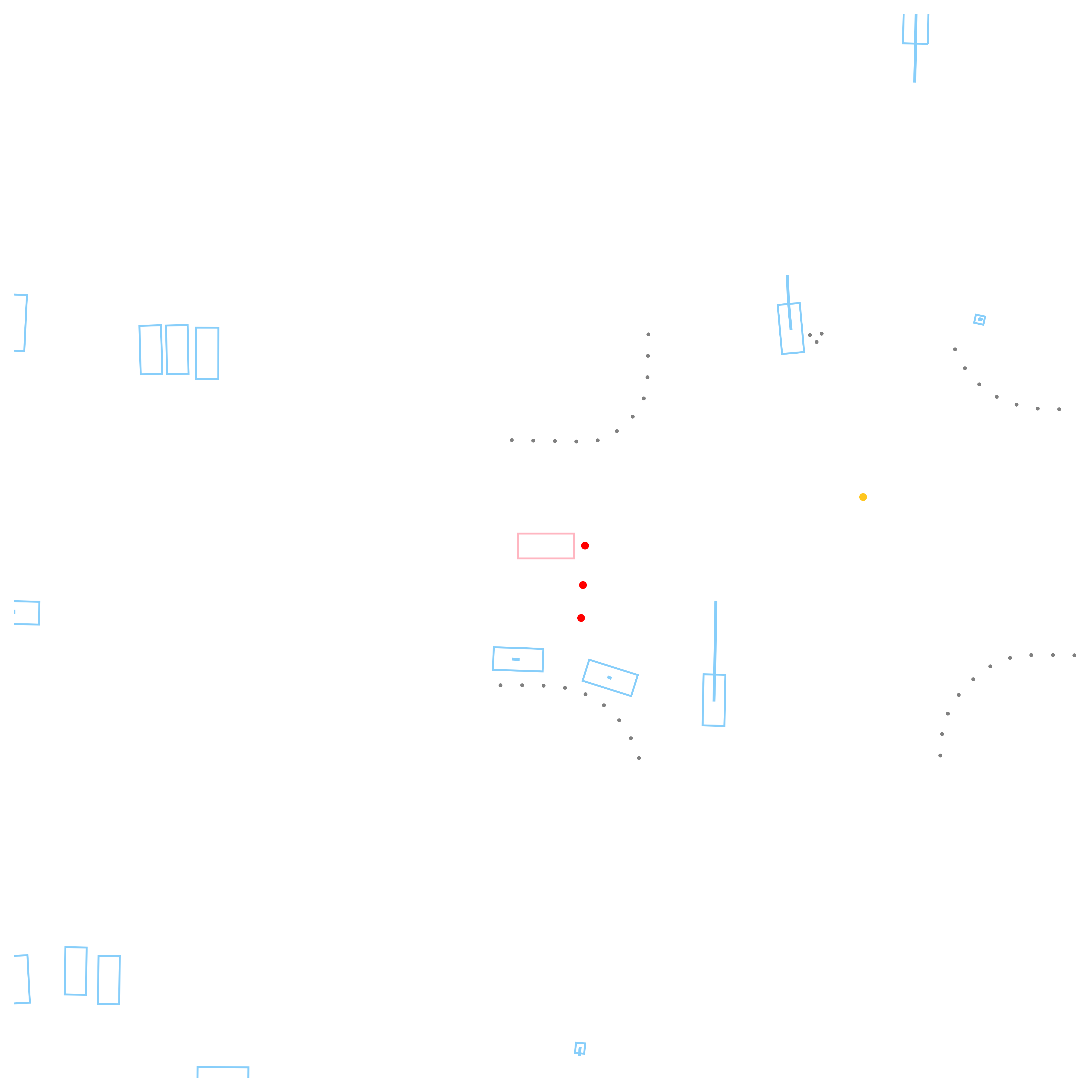}}
            \caption{road edge, goal point}
            \label{fig:base_obs}
        \end{subfigure}
        
        \begin{subfigure}[t]{\textwidth}
            \centering
            \fbox{\includegraphics[width=0.6\textwidth]{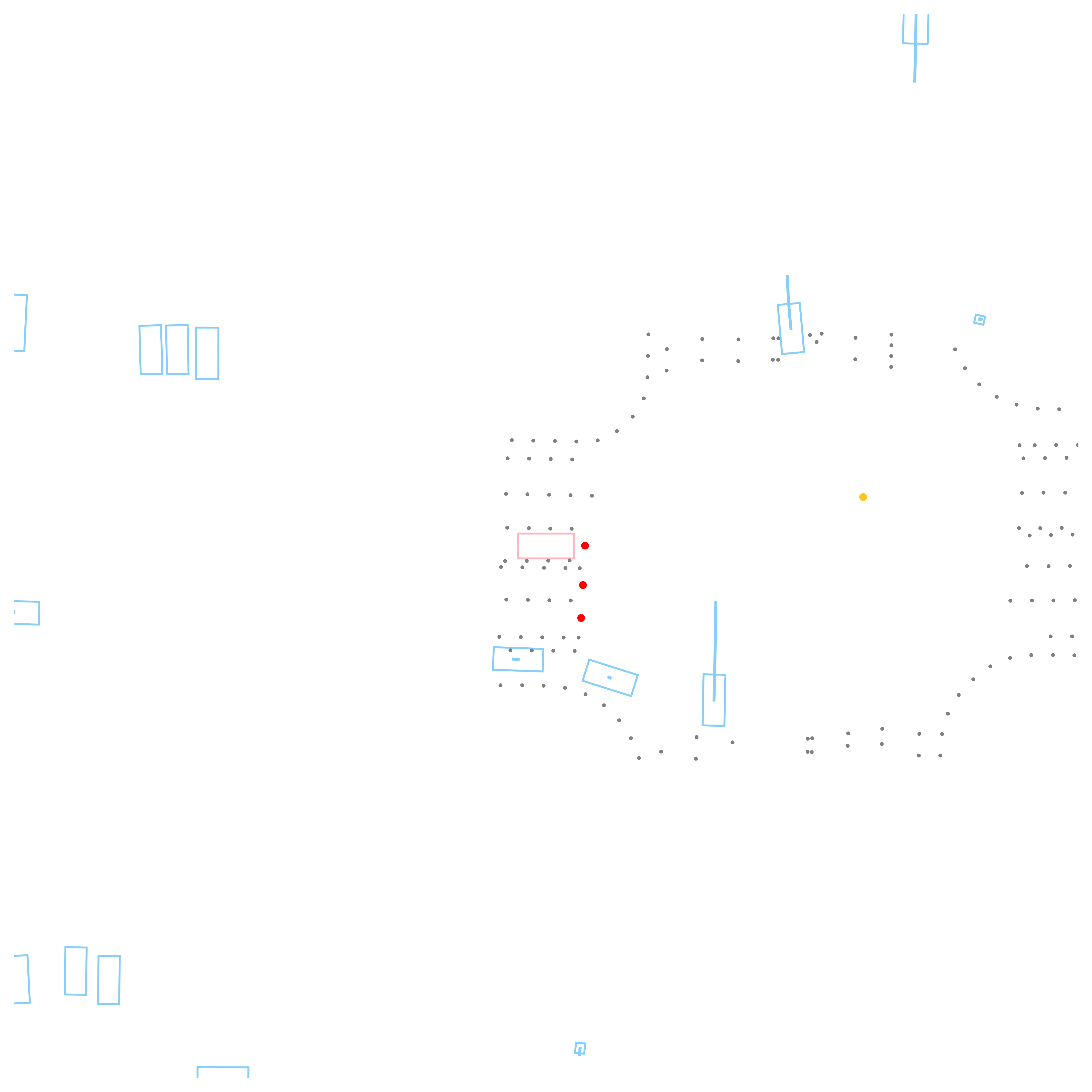}}
            \caption{road edge, road line}
            \label{fig:lane_obs}
        \end{subfigure}
        
        \begin{subfigure}[t]{\textwidth}
            \centering
            \fbox{\includegraphics[width=0.6\textwidth]{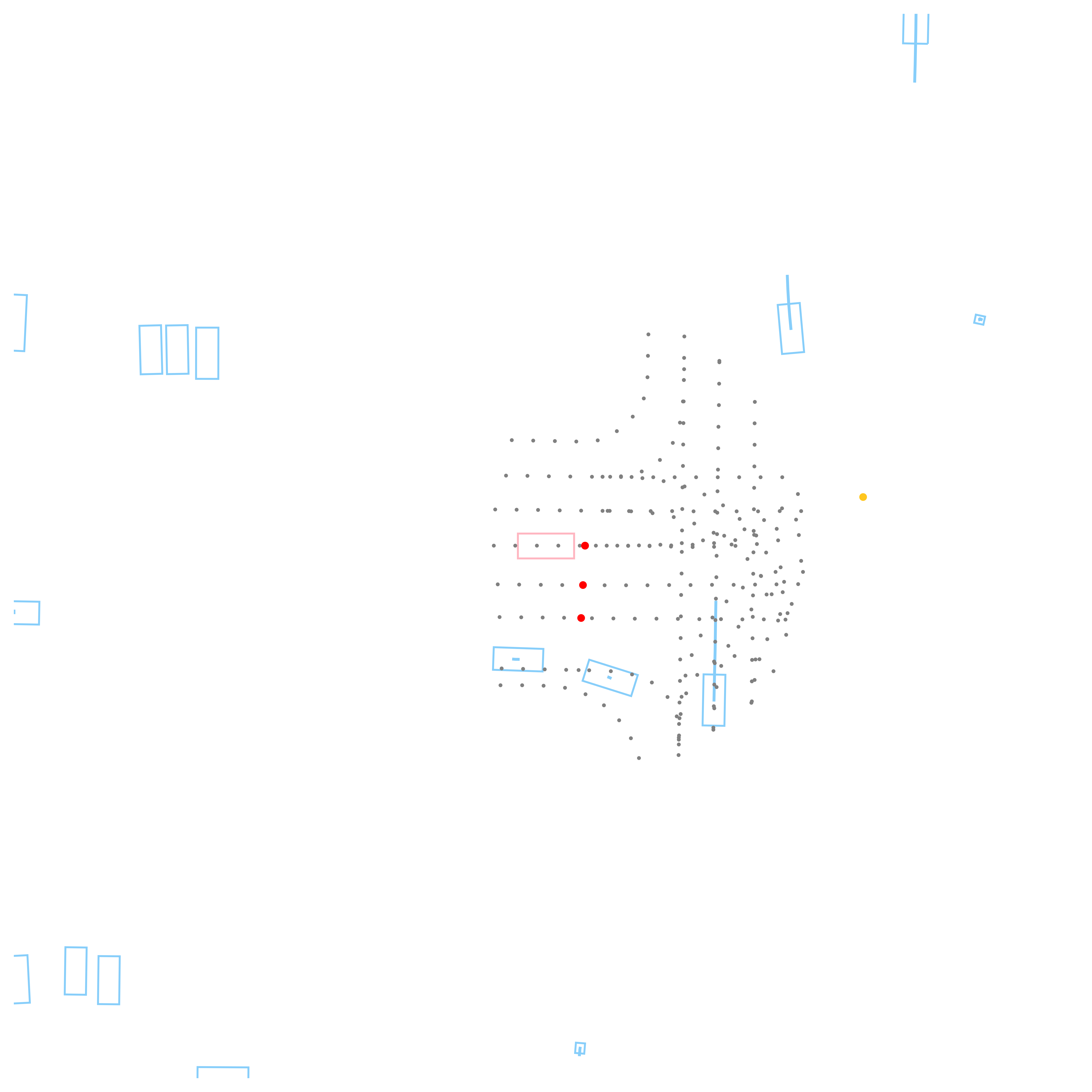}}
            \caption{road edge, lane center}
            \label{fig:road_obs_goal}
        \end{subfigure}
    \end{minipage}

\captionsetup{justification=centering}
\caption{Illustration of the observation functions tested in \autoref{sec:case_study}.}
\label{fig:observations_illustrated}
\end{figure}

\begin{table}[ht]
  \centering
  \caption{Observation configuration hyperparameter sweep results on WOMD Valid dataset. All experiments use SAC with LQ encoder and navigation reward (progression=0.2, off-route=-0.2).}
  \label{tab:observation_sweep}
  \small
  \begin{tabular}{cccc|c|cccc|cc}
    \toprule
    \multicolumn{4}{c|}{\textbf{Observation Config}} & \textbf{Road} & \multicolumn{4}{c|}{\textbf{Safety Metrics}} & \multicolumn{2}{c}{\textbf{Performance}} \\
    \textbf{Steps} & \textbf{Obj} & \textbf{Path} & \textbf{Gap} & \textbf{Features} & \textbf{Collision} & \textbf{Offroad} & \textbf{Red-light} & \textbf{Progress} & \textbf{Accuracy} & \textbf{V-Max} \\
    & & \textbf{Pts} & & & $\downarrow$ & $\downarrow$ & $\downarrow$ & $\uparrow$ & $\uparrow$ & $\uparrow$ \\
    \midrule
    1 & 8 & 1 & 30 & road\_edge & 1.54 & 0.62 & 0.27 & 165.65 & 97.11 & 0.81 \\
    1 & 8 & 1 & 30 & road\_edge+lane & 1.79 & 1.25 & 0.43 & 179.84 & 96.03 & 0.82 \\
    1 & 8 & 1 & 30 & road\_edge+line & 1.48 & 1.08 & 0.35 & 151.91 & 96.37 & 0.79 \\
    1 & 8 & 10 & 5 & road\_edge & 1.54 & 0.63 & 0.21 & 169.1 & 96.97 & 0.87 \\
    1 & 8 & 10 & 5 & road\_edge+lane & 1.46 & 1.31 & 0.15 & 173.9 & 96.54 & 0.84 \\
    1 & 8 & 10 & 5 & road\_edge+line & 1.41 & 0.92 & 0.14 & 161.0 & 96.92 & 0.84 \\
    1 & 16 & 1 & 30 & road\_edge & 1.56 & 0.73 & 0.20 & 163.91 & 96.80 & 0.79 \\
    1 & 16 & 1 & 30 & road\_edge+lane & 1.08 & 0.71 & 0.19 & 142.1 & 97.53 & 0.87 \\
    1 & 16 & 1 & 30 & road\_edge+line & 1.33 & 0.81 & 0.17 & 134.45 & 96.88 & 0.80 \\
    1 & 16 & 10 & 5 & road\_edge & 1.28 & 0.64 & 0.29 & 160.5 & 97.33 & 0.87 \\
    1 & 16 & 10 & 5 & road\_edge+lane & 1.60 & 0.87 & 0.15 & 161.9 & 96.55 & 0.87 \\
    1 & 16 & 10 & 5 & road\_edge+line & 1.61 & 0.86 & 0.25 & 168.06 & 96.66 & 0.84 \\
    \midrule
    3 & 8 & 1 & 30 & road\_edge & 1.38 & 0.61 & 0.15 & 161.93 & 97.30 & 0.81 \\
    3 & 8 & 1 & 30 & road\_edge+lane & 1.09 & 0.82 & 0.23 & 150.5 & 97.28 & 0.84 \\
    3 & 8 & 1 & 30 & road\_edge+line & 1.64 & 1.39 & 0.39 & 145.67 & 95.93 & 0.78 \\
    3 & 8 & 10 & 5 & road\_edge & 1.84 & 0.62 & 0.13 & 172.5 & 96.51 & 0.86 \\
    3 & 8 & 10 & 5 & road\_edge+lane & 1.92 & 1.33 & 0.26 & 174.5 & 96.05 & 0.85 \\
    3 & 8 & 10 & 5 & road\_edge+line & 1.23 & 0.84 & 0.35 & 170.8 & 97.13 & 0.84 \\
    3 & 16 & 1 & 30 & road\_edge & 1.00 & 1.06 & 0.17 & 154.22 & 97.10 & 0.81 \\
    3 & 16 & 1 & 30 & road\_edge+lane & 1.49 & 1.26 & 0.20 & 151.0 & 96.46 & 0.85 \\
    3 & 16 & 1 & 30 & road\_edge+line & 1.56 & 1.32 & 0.11 & 137.78 & 96.30 & 0.79 \\
    3 & 16 & 10 & 5 & road\_edge & 1.14 & 0.69 & 0.12 & 139.2 & 97.09 & 0.87 \\
    3 & 16 & 10 & 5 & road\_edge+lane & 1.03 & 0.94 & 0.10 & 142.2 & 97.18 & 0.87 \\
    3 & 16 & 10 & 5 & road\_edge+line & 1.17 & 0.85 & 0.22 & 152.5 & 97.12 & 0.85 \\
    \midrule
    5 & 8 & 1 & 30 & road\_edge & 1.70 & 0.68 & 0.28 & 167.38 & 96.82 & 0.80 \\
    5 & 8 & 1 & 30 & road\_edge+lane & 1.52 & 1.33 & 0.19 & 169.8 & 96.38 & 0.83 \\
    5 & 8 & 1 & 30 & road\_edge+line & 1.53 & 1.17 & 0.15 & 162.69 & 96.64 & 0.82 \\
    5 & 8 & 10 & 5 & road\_edge & 1.23 & 0.92 & 0.21 & 166.4 & 97.04 & 0.87 \\
    5 & 8 & 10 & 5 & road\_edge+lane & 1.59 & 0.87 & 0.15 & 155.25 & 96.78 & 0.81 \\
    5 & 8 & 10 & 5 & road\_edge+line & 1.26 & 1.07 & 0.15 & 144.97 & 96.74 & 0.83 \\
    5 & 16 & 1 & 30 & road\_edge & 1.35 & 0.62 & 0.25 & 158.57 & 97.26 & 0.82 \\
    5 & 16 & 1 & 30 & road\_edge+lane & 1.45 & 1.22 & 0.24 & 160.5 & 96.56 & 0.84 \\
    5 & 16 & 1 & 30 & road\_edge+line & 1.15 & 0.98 & 0.15 & 147.78 & 97.03 & 0.83 \\
    5 & 16 & 10 & 5 & road\_edge & 1.09 & 0.73 & 0.17 & 153.6 & 97.24 & \textbf{0.88} \\
    5 & 16 & 10 & 5 & road\_edge+lane & 1.55 & 1.20 & 0.18 & 181.9 & 96.49 & 0.84 \\
    5 & 16 & 10 & 5 & road\_edge+line & 1.34 & 1.09 & 0.20 & 153.5 & 96.80 & 0.86 \\
    \bottomrule
  \end{tabular}
      \label{tab:obs_sweep}
\end{table}

\clearpage

\section{Training hyperparameters}
\label{sec:hyperparameters}

We details hyperparameters used for all learning algorithms in the experiments. Every algorithm are trained in one single NVIDIA L4.

\begin{table}[ht]
    \centering
    \captionsetup{justification=centering} 
    \caption{Algorithms hyperparameters used in experiments}
    \begin{tabular}{@{}l|cccc@{}}
        \toprule
        \textbf{Parameter} & \textbf{BC} & \textbf{SAC} & \textbf{BC+SAC} & \textbf{PPO} \\
        \midrule
        Total Steps & 200M & 25M & 25M & 200M \\
        Num Envs & 16 & 16 & 16 & 16 \\
        Unroll Length & 80 & 1 & 1(RL)/80(IL) & 40 \\
        IL frequency & - & - & 8 & - \\
        Learning Rate & 1e-4 & 1e-4 & 1e-4/5e-5 & 1e-4 \\
        Batch Size & 64 & 64 & 64 & 512 \\
        Num Minibatches & - & - & - & 16 \\
        Discount & - & 0.99 & 0.99 & 0.99 \\
        Entropy $\alpha$ & - & 0.2 & 0.2 & 0.2 \\
        Updates/step & 32 & 4 & 8 & 4 \\
        Buffer size & - & 1000000 & 1000000 & - \\
        Learning start & - & 50000 & 50000 & - \\
        GAE factor & - & - & - & 0.95 \\
        Clip $\epsilon$ & - & - & - & 0.2 \\
        Loss function & Log\_prob & - & - & - \\
        \bottomrule
    \end{tabular}
    \label{tab:alg_hyperparameters}
\end{table}

\begin{table}[ht]
  \centering
  \caption{Reward shaping hyperparameter sweep results on WOMD Valid dataset. All experiments use SAC with LQ encoder, 5 observation steps, 8 objects, and road\_edge features.}
  \label{tab:reward_sweep}
  \small
  \begin{tabular}{cccc|cccc|cc}
    \toprule
    \multicolumn{4}{c|}{\textbf{Reward Weights}} & \multicolumn{4}{c|}{\textbf{Safety Metrics}} & \multicolumn{2}{c}{\textbf{Performance}} \\
    \textbf{Prog.} & \textbf{Off-route} & \textbf{Comfort} & \textbf{Overspeed} & \textbf{Collision} & \textbf{Offroad} & \textbf{Red-light} & \textbf{Progress} & \textbf{Accuracy} & \textbf{V-Max} \\
    & & & & $\downarrow$ & $\downarrow$ & $\downarrow$ & $\uparrow$ & $\uparrow$ & $\uparrow$ \\
    \midrule
    0.1 & -0.3 & 0 & 0 & 1.08 & 0.57 & 0.15 & 113.2 & 97.64 & 0.86 \\
    0.1 & -0.6 & 0 & 0 & 1.22 & 0.63 & 0.13 & 111.1 & 97.06 & 0.85 \\
    0.1 & -1.0 & 0 & 0 & 0.94 & 0.55 & 0.08 & 100.1 & 97.50 & 0.83 \\
    0.2 & -0.2 & 0 & 0 & 1.18 & 0.67 & 0.15 & 155.4 & 97.45 & 0.87 \\
    0.3 & -0.3 & 0 & 0 & 1.50 & 0.75 & 0.28 & 185.6 & 97.11 & 0.87 \\
    0.3 & -0.6 & 0 & 0 & 1.44 & 0.67 & 0.32 & 184.7 & 96.94 & 0.87 \\
    0.3 & -0.6 & 0 & 0 & 1.87 & 0.74 & 0.25 & 187.2 & 96.49 & 0.85 \\
    0.3 & -1.0 & 0 & 0 & 1.94 & 0.98 & 0.23 & 181.3 & 96.16 & 0.84 \\
    0.5 & -0.6 & 0 & 0 & 1.83 & 0.85 & 0.24 & 197.3 & 96.49 & 0.85 \\
    0.5 & -0.6 & 0 & 0 & 2.26 & 0.71 & 0.18 & 190.1 & 95.42 & 0.81 \\
    0.5 & -1.0 & 0 & 0 & 1.69 & 0.71 & 0.21 & 192.8 & 96.43 & 0.84 \\
    \midrule
    0.2 & -0.2 & 0.1 & -0.1 & 1.49 & 0.55 & 0.19 & 164.3 & 97.19 & 0.86 \\
    0.2 & -0.2 & 0.1 & -0.2 & 1.37 & 0.71 & 0.14 & 168.5 & 97.01 & 0.86 \\
    0.2 & -0.2 & 0.2 & -0.1 & 1.06 & 0.64 & 0.20 & 139.9 & 97.25 & \textbf{0.88} \\
    0.2 & -0.2 & 0.2 & -0.2 & 1.26 & 0.62 & 0.27 & 162.2 & 97.26 & \textbf{0.88} \\
    0.3 & -0.2 & 0.1 & -0.1 & 1.86 & 0.66 & 0.27 & 176.2 & 96.44 & 0.86 \\
    0.3 & -0.2 & 0.2 & -0.1 & 2.22 & 0.69 & 0.42 & 202.5 & 96.24 & 0.80 \\
    0.3 & -0.2 & 0.2 & -0.2 & 1.93 & 0.66 & 0.26 & 188.0 & 96.25 & 0.83 \\
    \bottomrule
  \end{tabular}
\end{table}

\begin{table}[ht]
    \centering
    \captionsetup{justification=centering} 
    \caption{Encoders and decoders hyperparameters used in experiments}
    \begin{tabular}{@{}lp{2.5cm}p{6.5cm}@{}}
        \toprule
        
        \multicolumn{3}{c}{MLP policy decoder} \\ \midrule
        Parameter & Value & Description \\ \midrule
        \verb|Layer sizes| & [256, 64, 32] & Number and size of layers \\
        \midrule

        \multicolumn{3}{c}{MLP value decoder} \\ \midrule
        \verb|Layer sizes| & [256, 64, 32] & Number and size of layers \\
        \midrule
        
        \multicolumn{3}{c}{LQH encoder} \\ \midrule
        \verb|Embedding sizes| & [256, 256] & Embedding layer sizes \\
        \verb|dk| & 64 & Dense encoder dimension \\
        \verb|num_latents| & 16 & Learnable latent size \\
        \verb|cross_num_heads| & 2 & Attention heads count \\
        \verb|cross_head_feat| & 16 & Features per attention head \\
        \verb|ff_mult| & 2 & Feedforward layer multiplier \\
        \midrule
        
        \multicolumn{3}{c}{LQ encoder} \\ \midrule
        \verb|Embedding sizes| & [256, 256] & Embedding layer sizes \\
        \verb|depth| & 4 & Attention layers count \\
        \verb|num_latents| & 16 & Learnable latent size \\
        \verb|num_self_heads| & 2 & Self-attention heads \\
        \verb|self_head_feat| & 16 & Features per self-attention \\
        \verb|cross_num_heads| & 2 & Cross-attention heads \\
        \verb|cross_head_feat| & 16 & Features per cross-attention \\
        \verb|ff_mult| & 2 & Feedforward layer multiplier \\
        \midrule

        \multicolumn{3}{c}{MTR encoder} \\ \midrule
        \verb|Embedding sizes| & [256, 256] & Embedding layer sizes \\
        \verb|dk| & 64 & Dense encoder dimension \\
        \verb|num_latents| & 16 & Learnable latent size \\
        \verb|num_self_heads| & 2 & Self-attention heads \\
        \verb|self_head_feat| & 16 & Features per self-attention \\
        \verb|ff_mult| & 2 & Feedforward layer multiplier \\
        \verb|k| & 8 & Nearest objects in attention \\
        \midrule

        \multicolumn{3}{c}{Wayformer encoder} \\ \midrule
        \verb|Embedding sizes| & [256, 256] & Embedding layer sizes \\
        \verb|dk| & 64 & Dense encoder dimension \\
        \verb|num_latents| & 16 & Learnable latent size \\
        \verb|num_self_heads| & 2 & Self-attention heads \\
        \verb|self_head_feat| & 16 & Features per self-attention \\
        \verb|depth| & 2 & Attention layers count \\
        \verb|ff_mult| & 2 & Feedforward layer multiplier \\
        \verb|fusion_type| & late & Fusion mechanism type \\
        \bottomrule
        
    \end{tabular}
    \label{tab:enc_hyperparameters}
\end{table}

\begin{figure}[ht]
    \centering
    \begin{subfigure}[b]{1\linewidth}
        \centering
        \fbox{\includegraphics[width=0.28\linewidth]{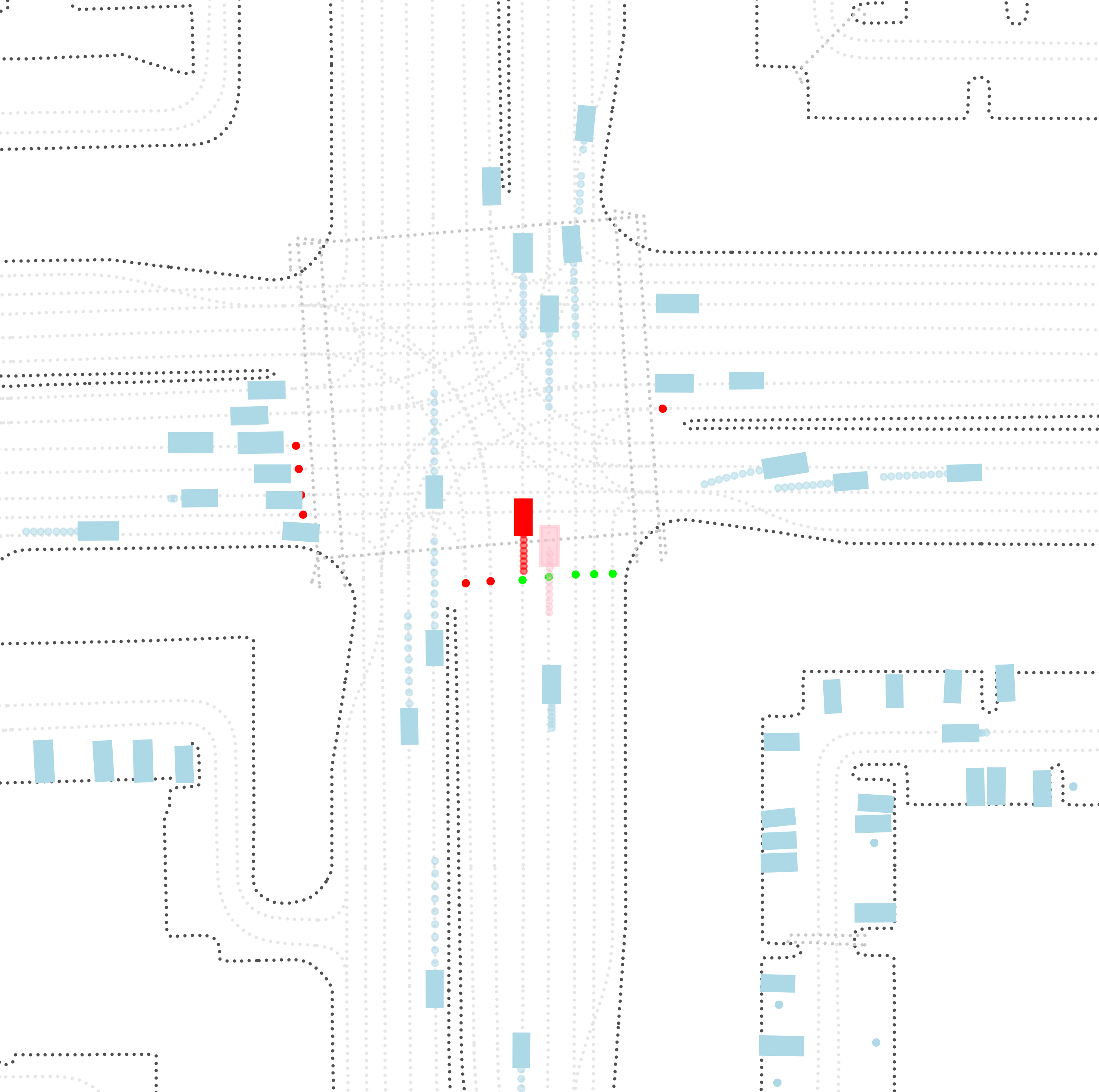}}
        \fbox{\includegraphics[width=0.28\linewidth]{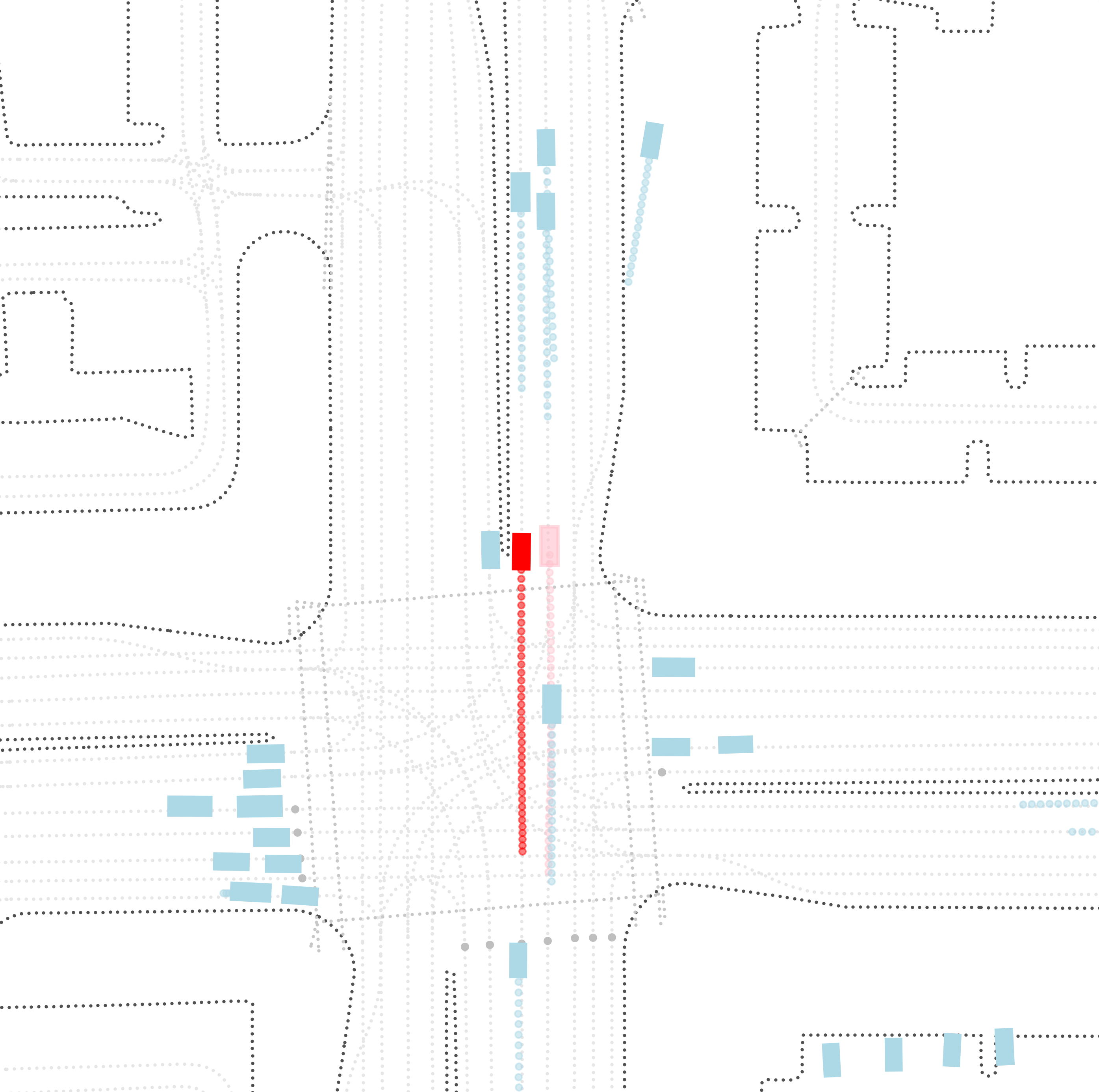}}
        \fbox{\includegraphics[width=0.28\linewidth]{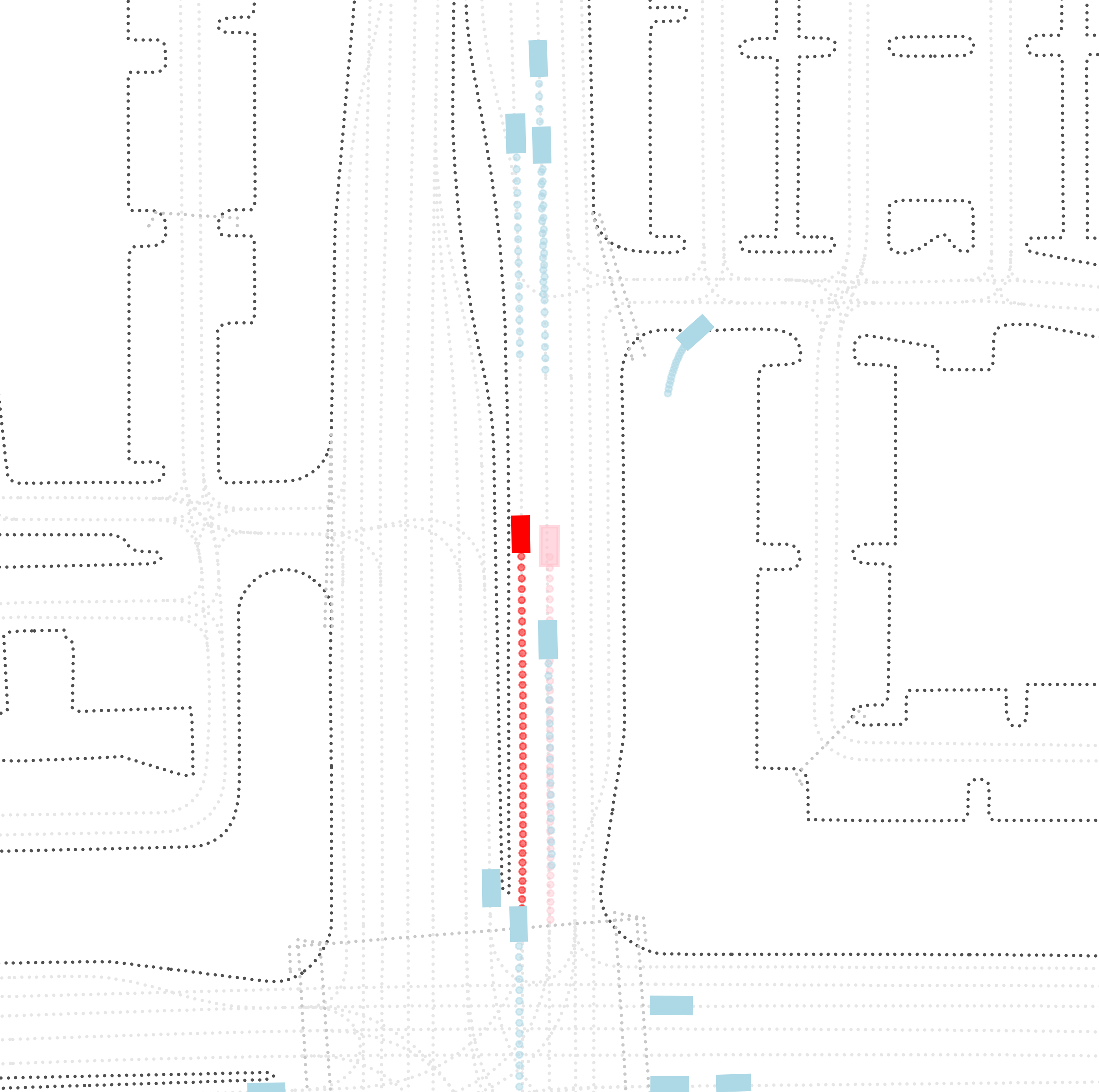}}
        \caption{Scene-centered view of the initial scenario.}
        \label{fig:no_attack}
    \end{subfigure}
    
    \begin{subfigure}[b]{1\linewidth}
        \centering
        \fbox{\includegraphics[width=0.28\linewidth]{figures//regents/init_scene.png}}
        \fbox{\includegraphics[width=0.28\linewidth]{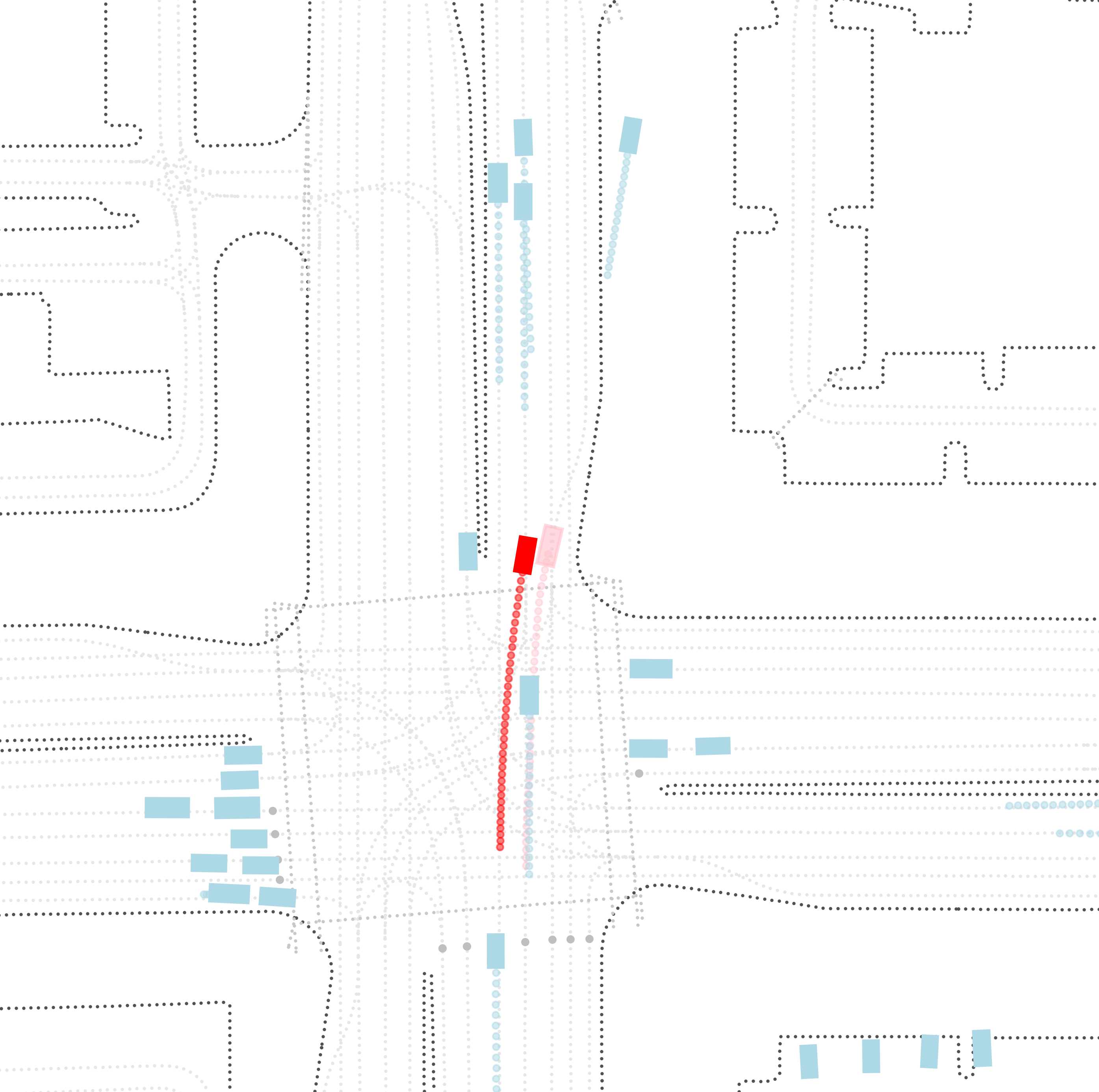}}
        \fbox{\includegraphics[width=0.28\linewidth]{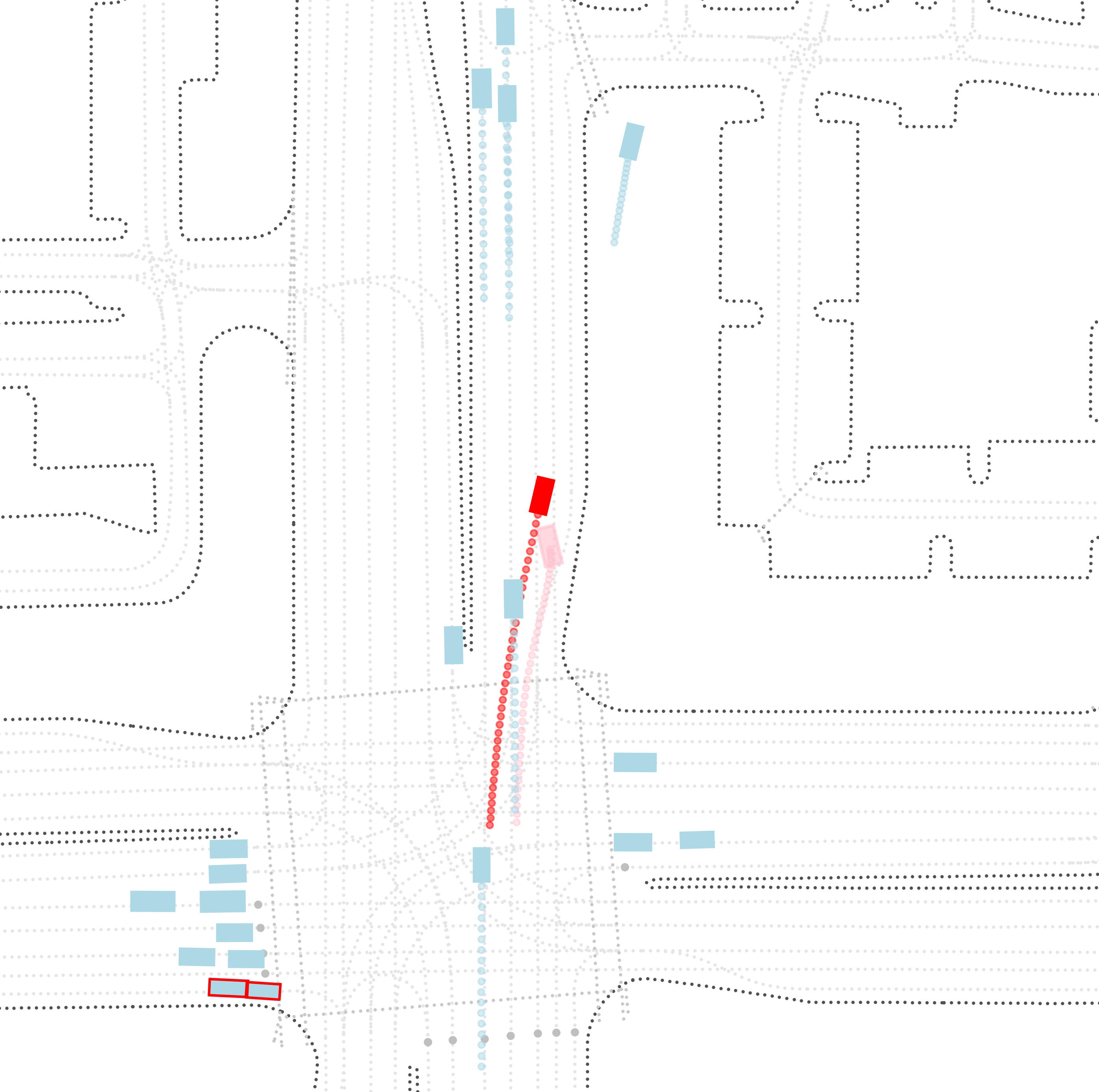}}
        \caption{Scene-centered view after applying ReGentS. The adversarial agent (in red) follows a trajectory designed to collide with the SDC agent (in pink). The RL agent adapts by deviating from its path to avoid the collision and then re-centering itself in the lane once the adversarial object has passed.}
        \label{fig:with_attack}
    \end{subfigure}

    \caption{Visualization of adversarial evaluation of our best RL agent using ReGentS \cite{yin_regents_2024}}
    \label{fig:regents_visualization}
\end{figure}

\end{document}